\documentclass{article}

\usepackage{PRIMEarxiv}

\usepackage[utf8]{inputenc} 
\usepackage[T1]{fontenc}    
\usepackage{hyperref}       
\usepackage{url}            
\usepackage{booktabs}       
\usepackage{amsfonts}       
\usepackage{nicefrac}       
\usepackage{microtype}      
\usepackage{lipsum}
\usepackage{fancyhdr}       
\usepackage{graphicx}       
\usepackage{subfigure}
\usepackage{amsmath}
\usepackage{algorithmic}
\usepackage[ruled,linesnumbered]{algorithm2e}

\graphicspath{{media/}}     

\title{SES: Bridging the Gap Between Explainability and Prediction of Graph Neural Networks
}

\author{
  Zhenhua Huang, Kunhao Li, Shaojie Wang, Zhaohong Jia \\
  School of Internet \\
  Anhui University \\
  Hefei\\
  \texttt{\{zhhuangscut, kunhomlihf, wsj.ahu\}@gmail.com, zhjia@mail.ustc.edu.cn} \\
   \And
  Wentao Zhu \\
  Amazon Research \\
  Seattle\\
  \texttt{wentaozhu91@gmail.com} \\
   \And
  Sharad Mehrotra \\
  University of California Irvine\\
  Irvine \\
  \texttt{sharad@ics.edu}\\
}

\begin{document}
\maketitle

\begin{abstract}
Despite the Graph Neural Networks' (GNNs) proficiency in analyzing graph data, achieving high-accuracy and interpretable predictions remains challenging. Existing GNN interpreters typically provide post-hoc explanations disjointed from GNNs' predictions, resulting in misrepresentations. Self-explainable GNNs offer built-in explanations during the training process. However, they cannot exploit the explanatory outcomes to augment prediction performance, and they fail to provide high-quality explanations of node features and require additional processes to generate explainable subgraphs, which is costly. To address the aforementioned limitations, we propose a self-explained and self-supervised graph neural network (SES) to bridge the gap between explainability and prediction. SES comprises two processes: explainable training and enhanced predictive learning. During explainable training, SES employs a global mask generator co-trained with a graph encoder and directly produces crucial structure and feature masks, reducing time consumption and providing node feature and subgraph explanations. In the enhanced predictive learning phase, mask-based positive-negative pairs are constructed utilizing the explanations to compute a triplet loss and enhance the node representations by contrastive learning.  

Extensive experiments demonstrate the superiority of SES on multiple datasets and tasks. SES outperforms baselines on real-world node classification datasets by notable margins of up to 2.59\% and achieves state-of-the-art (SOTA) performance in explanation tasks on synthetic datasets with improvements of up to 3.0\%. Moreover, SES delivers more coherent explanations on real-world datasets, has a fourfold increase in Fidelity+ score for explanation quality, and demonstrates faster training and explanation generating times. To our knowledge, SES is a pioneering GNN to achieve SOTA performance on both explanation and prediction tasks.
\end{abstract}


\section{Introduction}
\label{introduction}
Graph neural networks (GNNs) have become pivotal in handling graph data, proving essential in a variety of applications including node classification~\cite{metagnn2019, splitgnn2023}, knowledge representation~\cite{knowledge2021,knowledgerepre2023}, molecular classification~\cite{molecular2018, modular2022}, traffic prediction~\cite{traffic2021, traffic2022}, recommendation system~\cite{recsys2018, recsys2023}, sentiment analysis~\cite{senanalys2022, senanalys2_2022}, pose estimation~\cite{action2021, posest2023}, and text classification~\cite{textc2019, textc2022},~\textit{etc}.

One group of existing research on graph neural networks focuses on developing novel architectures to enhance predictive accuracy. Typical GNNs include graph convolution networks (GCN)~\cite{gcn2017}, graph attention networks (GAT)~\cite{gat2018}, GraphSAGE~\cite{graphsage2017}, graph isomorphism network (GIN)~\cite{gin2018}, ARMA~\cite{ARMA2021}, UniMP~\cite{transformconv2021}, FusedGAT~\cite{fusedgat2022}, and ASDGN~\cite{anti-symmetric2023},~\textit{etc}. However, these advancements have not adequately addressed the need for explainability in the representations learned by GNNs.

To address this, instance-level or model-level approaches have been proposed to offer explanations of GNNs, which are mostly post-hoc models. A post-hoc model is a statistical or predictive model constructed after data processing, enabling retrospective analysis and interpretability of variable relationships~\cite{posthoc2010}. Noteworthy instance-level post-hoc GNN explainers include GNNExplainer~\cite{gnnexplainer2019}, PGExplainer~\cite{pgexplainer2020}, PGMExplainer~\cite{pgmexplainer2020}, and GraphLIME~\cite{graphlime2022},~\textit{etc}. instance-level models offer individualized explanations of structures or features while model-level post-hoc explanation models offer abstract subgraphs conducive to classification for a specific GNN model, such as XGNN~\cite{xgnn2020}, PAGE~\cite{page2022} and GNNInterpreter~\cite{gnninterpreter2023},~\textit{etc}. The post-hoc explainers contribute to understanding the inner workings of GNNs. Due to the sequential nature of post-hoc GNN explainers, which explain one node at a time, interpreting a batch of nodes using the aforementioned approaches can be time-consuming. Moreover, they require an auxiliary model to explain a target GNN leading to potential bias and misrepresentations~\cite{segnn2021}.

\begin{figure}[h]
  \centering
        \includegraphics[scale=0.4]{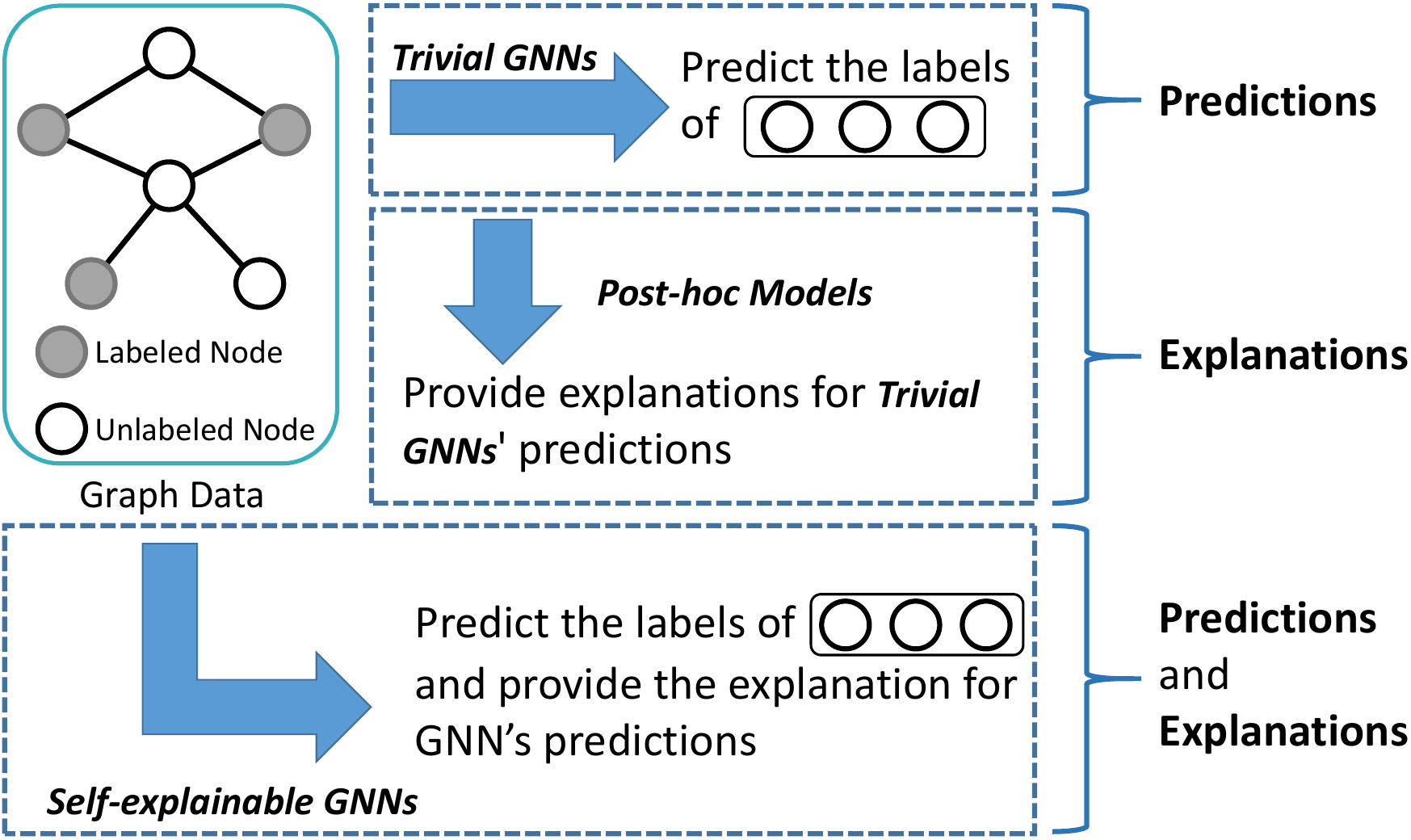}
  \caption{Categories of GNNs from explanation and prediction perspectives.}
  \label{cate_fig}
\end{figure}

\begin{table*}[!ht]
    \caption{Characteristics of SES and other types of Graph Neural Networks.}
    \centering
    \scriptsize
    \begin{tabular}{c|ccc}
    \toprule
        Method & Prediction & Explainable & Improvement \\ \midrule
        Trivial graph neural networks (GCN~\cite{gcn2017}, GAT~\cite{gat2018}, GIN~\cite{gin2018}, etc.) & \checkmark &  & \\
        Post-hoc explanation models (GNNExplainer~\cite{gnnexplainer2019}, PGExplainer~\cite{pgexplainer2020}, PGMExplainer~\cite{pgmexplainer2020}, etc.) &  & \checkmark &  \\
        Self-explainable models (SEGNN~\cite{segnn2021}, ProtGNN~\cite{protgnn2022}, PxGNN~\cite{pxgnn2022}, etc.) & \checkmark & \checkmark & \\
        SES (Ours) & \checkmark & \checkmark & \checkmark \\\bottomrule
    \end{tabular}
    \label{method_cate}
    \vspace{-0.5cm}
\end{table*}

In recent years, self-explainable GNN models have been introduced to address the limitations of post-hoc models, including SEGNN~\cite{segnn2021}, ProtGNN~\cite{protgnn2022}, and PxGNN~\cite{pxgnn2022}. SEGNN achieves explainable node classification by identifying the K-nearest labeled nodes for each unlabeled node by leveraging node similarity and local structure similarity. However, computing similarity scores between the test node and all labeled nodes is computationally expensive, and finding K-nearest labeled nodes by similarities is unstable. ProtGNN merges prototype learning with GNNs to provide explanations derived from case-based reasoning. However, the inclusion of prototype reflection and Monte Carlo tree search brings additional computational costs. Moreover, the node prototype at the cluster boundary in ProtGNN is not as distinguishable as the subgraph structure, which compromises its node classification efficacy. PxGNN~\cite{pxgnn2022} provides class-level explanations and makes classifications by a prototype generator constraining on learnable prototype embeddings. It relies on the prototype generator's capacity and non-representative prototypes can result in subpar interpretative and predictive performance.

Self-explainable GNNs aim to enhance interpretability by jointly modeling representations and explanations. Nevertheless, current self-explainable GNNs exhibit the following issues: (1) Unlike post-hoc methods, they fail to indicate the importance of node features, making them difficult to apply to a wider range of explanation tasks. (2) These methods require additional computations overhead to obtain or search for explanatory subgraphs~\cite{segnn2021, protgnn2022}, which is costly. (3) Empirical evidence suggests that these methods exhibit unsatisfactory predictive performance, resulting from ineffective utilization of the interpretation outcomes. Consequently, a chasm persists between the predictive capabilities and the explanations provided by GNNs, which hampers their utility in downstream applications.

As illustrated in Fig.~\ref{cate_fig}, trivial GNNs are typically limited to prediction tasks, offering no interpretive insights. Post-hoc explanation models create explanations for graph data by trained GNNs. In contrast, self-explainable GNNs provide explanations concurrently with the training of node representations. However, they fail to enhance prediction accuracy. We argue that cogent explanations deepen the understanding of graphs and bolster the model's training efficacy and prediction accuracy. To overcome the shortcomings inherent in existing GNNs, we propose a \underline{S}elf-\underline{E}xplained and self-\underline{S}upervised model (SES) to bridge the gap between the explainability and prediction accuracy of GNN.  SES harmonizes the interpretation generation and training of GNNs, instead of treating them as separate processes. The characteristics of GNNs are summarized in TABLE~\ref{method_cate}. 

The SES approach consists of two phases: Explainable Training and Enhanced Predictive Learning. During the Explainable Training phase, SES concurrently optimizes the mask generator with the graph encoder to ensure that the feature and structure mask align with the aggregation performed by the backbone GNN. It eliminates the need for computationally expensive searches or constructions of explanatory subgraphs, which significantly reduces the time consumption. The masks are co-trained with the graph encoder with a dedicated loss allowing it to capture the most relevant features and substructures for each node and resulting in accurate explanations. Unlike existing self-explainable GNNs, SES generates masks that provide both feature and structure explanations, making it suitable for a wider range of explanation tasks.

To utilize these explanations and improve prediction performance, inspired by contrastive learning (a self-supervised method), mask-based positive and negative pairs are created to compute a triplet loss that ensures nodes with similar structures are closely situated and discriminates nodes with dissimilar structural attributes. Simultaneously, supervised learning steers the self-supervised training to ensure its alignment with the prediction task. We refer to the second phase as "Enhanced Predictive Learning" as it enhances the effects of self-supervised training and prevents deviation from the prediction task. Experimental results demonstrate the efficiency and effectiveness of SES across multiple tasks. In summary, our contributions can be summarized as follows:
\begin{itemize}
    \item We investigate the limitations of current self-explainable GNNs: inadequate feature explanations, costly computations, and plain prediction performance.
    \item We propose a pioneering self-explained and self-supervised graph neural network that bridges gaps between explainability and prediction of GNNs.
    \item SES provides more reasonable interpretations of models and significantly reduces the time consumption of generating explanations.
    \item Extensive quantitative experiments demonstrate that SES achieves SOTA performance in both explanation and prediction tasks on real-world and synthetic datasets.
\end{itemize}

\section{Related Works}
\subsection{Prediction of Graph Neural Networks}
Graph neural networks (GNNs) have shown a strong representation ability for dealing with graph format data~\cite{gnnsurvey2021}. The prediction of GNNs is to utilize nodes' or graphs' representation to classify their labels for the downstream tasks. GCN~\cite{gcn2017} is a typical GNN based on the message-passing among neighbors to aggregate information. GAT~\cite{gat2018} utilizes an attention mechanism to enhance the ability of information aggregation based on GCN. GraphSAGE~\cite{graphsage2017} applies a new sampling method to aggregate neighbor nodes to make GNN more scalable and effective. ARMA~\cite{ARMA2021} is a graph convolutional layer by autoregressive and moving average filters to provide a more flexible frequency response and better global graph structure representation. FusedGAT~\cite{fusedgat2022} is an optimized version of GAT that fuses message-passing computation for accelerated execution and lower memory footprint. ASDGN~\cite{anti-symmetric2023} is a stable and non-dissipative DGN design framework conceived through ordinary differential equations and preserves remote information between nodes. RAHG~\cite{rahg2023} is a role-aware GNN considering role features to improve node representation. These GNNs focus on improving the prediction performance without providing explanations.

\subsection{Explanation of Graph Neural Networks}
To offer interpretations of GNNs, a considerable amount of methods or GNN explainers are proposed~\cite{gnnexplainsurvey2022}. Many popular explainers are post-hoc models that can be categorized into instance-level and model-level approaches. GNNExplainer~\cite{gnnexplainer2019} is the pioneering instance-level model that provides edge and feature explanations by maximizing the mutual information between the prediction of a GNN and the distribution of possible subgraph structures. PGExplainer~\cite{pgexplainer2020} adopts a deep neural network to parameterize the explanation generation process, enabling it to provide multi-instance explanations naturally. GraphLIME~\cite{graphlime2022} is a local interpretable model that explains graphs using the Hilbert-Schmidt Independence Criterion (HSIC) Lasso, which focuses on providing feature explanations for graph data. XGNN~\cite{xgnn2020} is the first model-level approach to explain GNNs. It accomplishes this by training a graph generator that maximizes specific predictions made by a model. However, these post-hoc explainers for GNNs rely on trained models and require an additional model or process to support explanations. This dependency leads to a potential misunderstanding between the explainable models and GNNs.

To address the limitations of post-hoc explanations, some self-explainable GNNs that provide explanations during the training process are proposed. SEGNN~\cite{segnn2021} is the first self-explainable GNN, which utilizes an interpretable similarity module to find the K-nearest labeled nodes for each unlabeled node, enabling explainable node classification. The module considers node and local structure similarity to identify the nearest labeled nodes. ProtGNN~\cite{protgnn2022} combines prototype learning with GNNs, offering a new perspective on GNN explanations. The explanations provided by ProtGNN are derived from the case-based reasoning process. PxGNN ~\cite{pxgnn2022} is a prototype-based self-explainable GNN that can simultaneously give accurate predictions and prototype-based explanations of predictions.

However, results suggest that SEGNN requires substantial memory and ProtGNN is computationally expensive. Additionally, despite their strengths in elucidating the decision-making processes of GNNs and providing explanations, these models fail to achieve classification accuracy on par with trivial GNNs.

\subsection{Self-Supervised Learning of Graph Neural Networks}
Self-supervised learning has emerged as a promising technique for training deep learning models without extensive labeled data~\cite{self_supervised_survey2020}. In GNNs, self-supervised learning has gained significant attention due to its ability to learn useful representations from graph data without extra labeling works~\cite{graphsupervised2022}. GraphCL~\cite{graphcl2020} develops comparative learning with data enhancement for GNN pre-training to address the heterogeneity of graph data. GCC~\cite{gcc2020} is an unsupervised graph representation learning framework that captures common network topology properties from multiple networks. MERIT~\cite{merit2021} combines the advantages of Siamese knowledge distillation and conventional graph contrastive learning.  HeCo~\cite{heco2021} contrasts heterogeneous graphs using two perspectives (network schema and meta-path) and trains an encoder to maximize the mutual information between node embeddings. MolCLR~\cite{molclr2022} is a self-supervised GNN framework for molecular contrast learning, proposing enhancement methods to ensure consistency within the same molecule and inconsistency among different molecules. These methods are effective in the pre-training and general training phases of GNNs. However, there are currently no approaches to integrate self-supervised learning into explainable GNNs training effectively.

\begin{table}[hb]
    \renewcommand\arraystretch{1.0}
    \centering
    \caption{Notations in SES.}
        \begin{tabular}{c|c}
            \toprule
            Symbols & Definition and description \\\hline
            G & A general graph \\
            $V$ & Node set \\
            $A$ & Adjacency matrix \\
            $A^{(k)}$ & The k-hop adjacency matrix of $A$ \\
            $\tilde{A}^{(k)}$ & The complement adjacency of $A^{(k)}$\\\hline
            $Z$ & The output of graph encoder\\
            $Z_m$ & The optimized output by masks of graph encoder \\
            $\hat{Z}$ & The output of graph encoder in enhanced predictive learning\\\hline
            $P_r$ & Relational neighbor set\\
            $P_n$ & Negative neighbor set \\\hline
            $\theta_e$ & The learnable parameters in graph encoder\\
            $\theta_m$ & The learnable parameters in mask generator\\\hline
            $M_f$ & The feature mask from SES\\
            $M_s$ & The structure mask from SES\\
            $\hat{M}_s$ & The transferred structure mask from SES\\
            $E_{feat}$ & The feature explanation based on $M_f$\\
            $E_{sub}$ & The substructure explanation based on $M_s$\\\hline
            $Y$ & The label set of nodes \\
            $\mathcal{L}_{xent}$ & The cross-entropy loss\\
            $\mathcal{L}_{sub}$ & The subgraph loss\\
            $\mathcal{L}_{triplet}$ & The triplet loss\\
            \bottomrule
        \end{tabular}
        \label{notations}
        \vspace{-0.3cm}
\end{table}

\section{Problem Definition}
\begin{figure*}[h]
  \centering
        \includegraphics[scale=0.45]{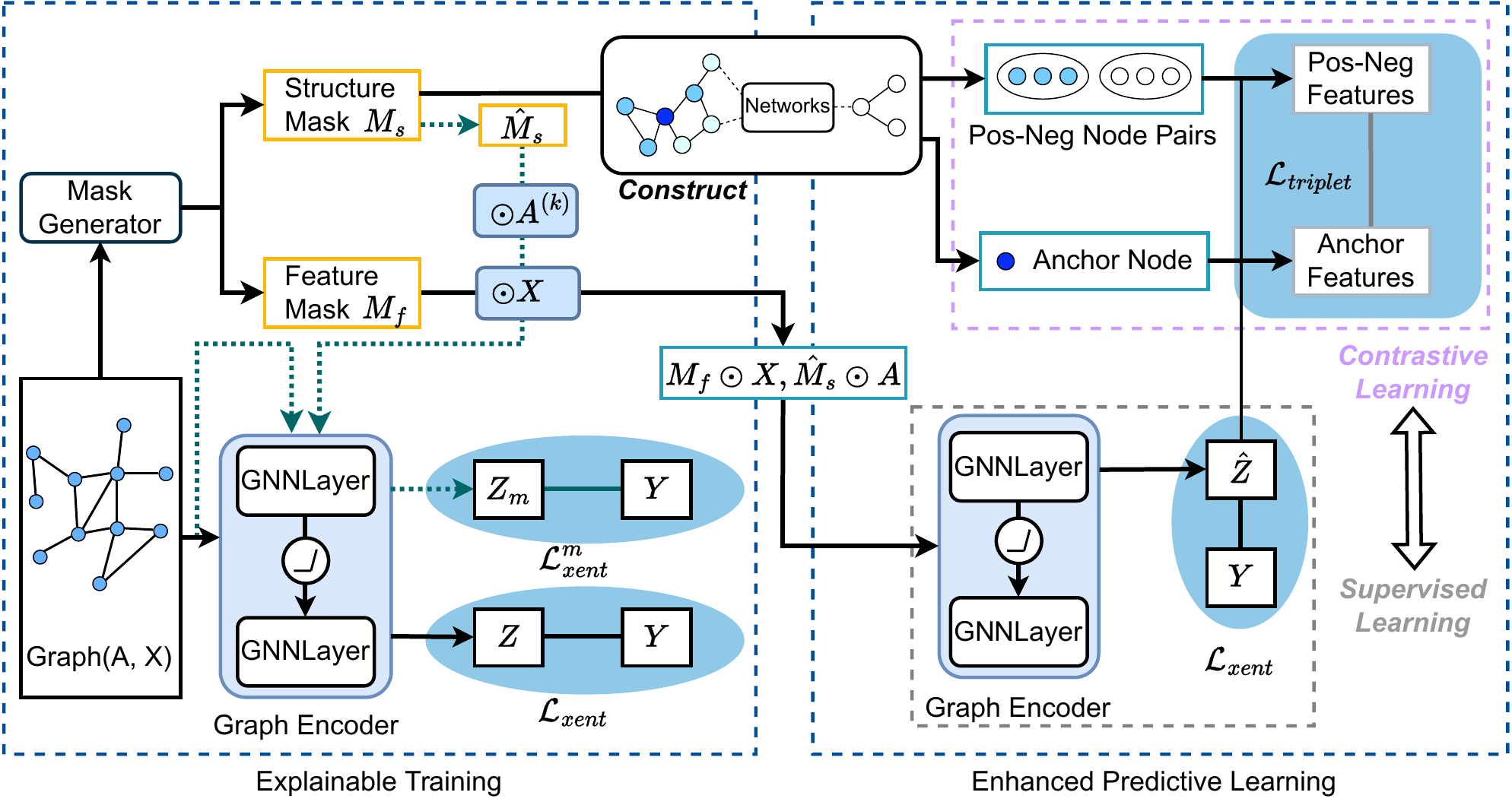}
  \caption{The framework of SES involves two phases: explainable training and enhanced predictive learning. The parameters of the graph encoder are shared in two phases. The green dashed arrows marked the process where the mask generator and the graph data are fed into the graph encoder.}
  \label{framework}
  \vspace{-0.2cm}
\end{figure*}

A graph is denoted by $G = (V, A, X)$, where $V = \{ v_1, \cdots, v_N\}$ denotes the node set with $N$ nodes, $A \in \mathcal{R}^{N \times N}$ denotes the adjacency matrix of $G$, $a_{ij} \in A$ and $a_{ij} = 1$ if $v_i$ and $v_j$ are connected. $X \in \mathcal{R}^{N \times F}$ represents the node features with $F$ dimensions. 

For predicting the node's label task (i.e., node classification task), the label set of nodes is denoted as $Y_L$. The labeled and unlabeled node sets from $V$ can be written as $V_L = \{v_{l1}, \cdots, v_{ln}\}$ and $V_U = \{v_{u1}, \cdots, v_{un}\}$. The objective of the node classification task is to utilize the graph $G$ and the set of labeled nodes $V_L$ to train the GNN, enabling the prediction of labels for the unlabeled nodes in $V_U$.

For the instance-level explanation, a model offers explanations for node features or substructures. A feature explanation in SES represented as $E_{feat} \in \mathcal{R}^{N \times F}$ is provided to clarify the significance of each feature dimension in predicting a node's label. Additionally, a substructure explanation denoted as $E_{sub} \in \mathcal{R}^{N \times N}$ in SES is applied to highlight the importance of a node's local neighbors.


Following GNNExplainer~\cite{gnnexplainer2019}, the problem of describing a self-explainable and self-supervised GNN can be written as:

Given a general graph $G(V, A, X)$, with labeled and unlabeled node set $V_L$ and $V_U$, we learn a self-explainable and discriminative self-supervised GNN $f$: $v_{ui} \rightarrow y$, which provide feature explanation $E_{feat}$ and substructure explanation $E_{sub}$ as the instance-level explanation for each $v_{ui} \in V_U$.

To facilitate comprehension, symbols used in this paper and their definitions are summarized in TABLE~\ref{notations}.

\section{Proposed Method}
The framework of SES is depicted in Fig.~\ref{framework}. SES consists of two primary phases: explainable training and enhanced predictive learning. 
SES employs a mask generator co-trained with a graph encoder in the explainable training phase. The mask generator generates feature masks for nodes and structure masks for the subgraphs and is optimized during training. In the enhanced predictive learning phase, SES constructs positive-negative node pairs based on the structure masks. These node pairs are then utilized to optimize the node features, which are subsequently processed by a shared graph encoder. By contrastive learning, SES designs a triplet loss to supervise the representation learning of nodes. Supervised learning is also employed to bolster the contrastive learning. The detailed procedures of SES is delineated in Algorithm \ref{SES_pseudocode}.

\subsection{Graph Encoder and Mask Generator}
\subsubsection{Graph Encoder}
We utilize a GNN backbone as a graph encoder to generate node embeddings. The GNN backbone is flexible and can be GCN~\cite{gcn2017} and GAT~\cite{gat2018}, etc. A general process of one convolution layer $Conv$ in GNN based on message passing~\cite{messagepassing2017} for a node $v$ is described as follows:
\begin{equation}
    h_v^{(l)}=COB^{(l)}\{AGG^{(l)}(h_u^{(l-1)}:u \in \mathcal{N}(v)),\ h_v^{(l-1)}\},
    \label{gnn_agg}
\end{equation}
where $h_v^{(l)}$ is the feature vector of $v$ in the $l^{(th)}$ layer and $\mathcal{N}(v)$ is the neighbors of $v$. $h_v^{(l-1)}$ represents the feature vector of $v$ in the $(l-1)^{(th)}$ layer.  $AGG$ is the function to aggregate features of nodes with their neighbors. $COB$ is the combination function to update the representation of nodes.

The process of a graph encoder with two convolution layers in Eq. (\ref{gnn_agg}) is summarized as follows:
\begin{equation}
    Z = Conv_2(\sigma(Conv_1(A,X)),A),
    \label{ge}
\end{equation}
where $\sigma$ is the activation function, and $Z$ is the output of the graph encoder used for node classification. $H \in \mathcal{R}^{N \times F_{fid}}=Conv_1(A, X))$ is the output from the first convolution layer $Conv_1$ employed for feature mask generation. $F_{hid}$ denotes the hidden size of $Conv_1$. $X=\{h^0_v, v \in V\}$, $H=\{h^1_v, v \in V\}$, $Z=\{h^2_v, v \in V\}$. 

\subsubsection{Mask Generator}
\begin{figure}[h]
  \centering
        \includegraphics[scale=0.5]{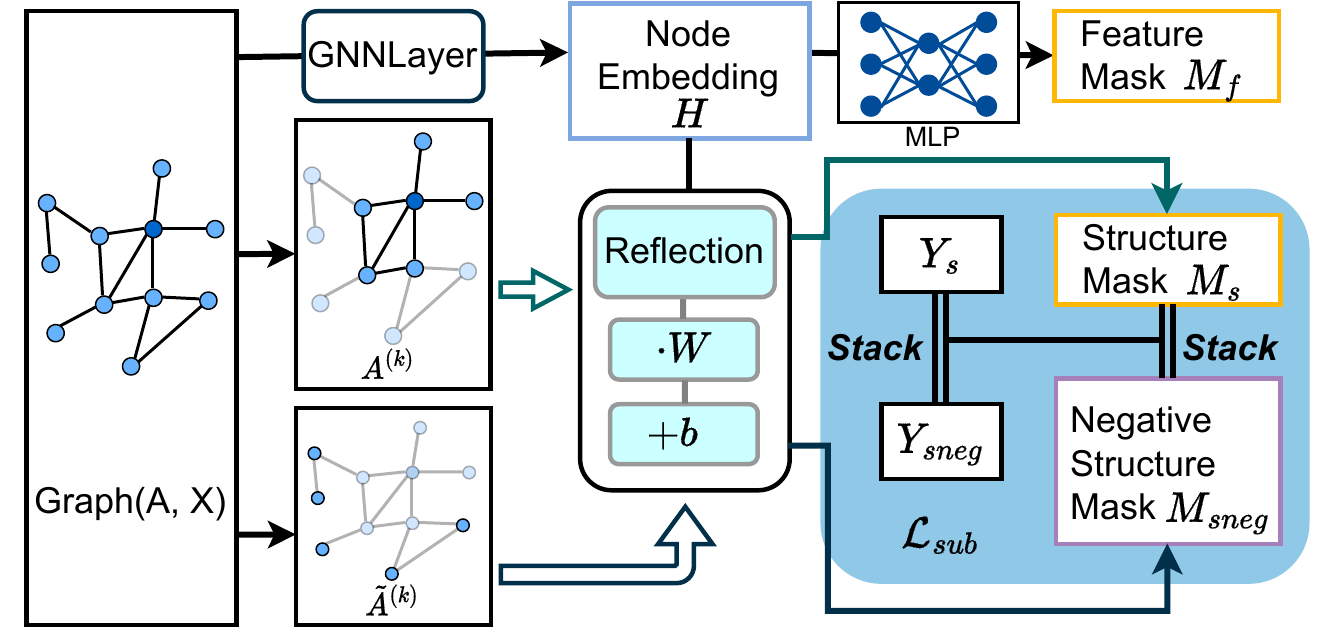}
  \caption{The framework of the mask generator in SES.}
  \label{maskgen}
\end{figure}

The masks are weight matrices to provide explanations of features and structures, emphasizing crucial feature dimensions and neighboring nodes. The framework of the global mask generator is depicted in Fig.~\ref{maskgen}. 

$H$ is passed through a multi-layer perception to generate a node feature mask denoted as $M_f$:
\begin{equation}
    M_f = \text{$MLP$}(H),
    \label{mf}
\end{equation}

To generate the structure mask $M_s$, a k-hop adjacency matrix denoted as $A^{(k)}$ is constructed. This matrix captures the neighbor relations of central nodes within their subgraphs. The structure mask $M_s \in \mathcal{R}^{N_k \times 1}$ is a composite matrix that combines output features of neighboring nodes and central nodes, $N_k$ is the number of edges in $A^{(k)}$.

To enhance the learning of the substructure, we conduct negative sampling on nodes that are not part of the subgraph of the central node and with different labels. The negative structure mask based on negative samples are represented as $M_{sneg} \in \mathcal{R}^{N_k \times 1}$, which is aligned with $M_s$ to compute the substructure loss in Eq. (\ref {Lsub}). $M_s$ and $M_{sneg}$ are constructed by the following formulations:
\begin{equation}
    \begin{aligned}
        &P_r(v_i)=\{j\mid a_{ij} \ne 0\}, \quad a_{ij} \in A^{(k)},\\
        &M_{s} = \sigma[W \cdot(\mathop{stk}\limits_{i=1}^N \mathop{cat}\limits_{k \in P_r(v_i)}(h_i,h_k))+b],\\
        &M_{sneg} = \sigma[W \cdot(\mathop{stk}\limits_{i=1}^N \mathop{cat}\limits_{k \in P_n(v_i)}(h_i,h_k))+b],
    \end{aligned}
    \label{ms}
\end{equation}
where $h_i = h^1_i$ is the output feature vector of node $i$'s from the first convolution layer $Conv_1$, and $h_k=h^1_k$ is the output feature vector of node $k$ but from $P_r(v_i)$ or $P_n(v_i)$. $P_r(v_i)$ represents the neighbor relational set of node $v_i$, while $P_n(v_i)$ denotes the negative set of node $v_i$. $P_n(v_i)$ is obtained by randomly selecting an equal number of $v_i$'s k-hop neighbors from the complement of the adjacency matrix $\tilde{A}^{(k)}$, where $\tilde{A}^{(k)} = I - A^{(k)}$. The function $cat$ performs concatenation operations on the $h_i$ and $h_k$, $stk$ is the column operation for stacking the result of $h_i$ and $h_k$'s concatenation. The concatenation function performs a concatenation operation along the first dimension of $h_i$ and $h_k$, while stacking is used to stack the concatenated results of $h_i$ and $h_k$ column-wise. Additionally, $W$ and $b$ represent the shared learnable weight and bias applied to $M_s$ and $M_{sneg}$, respectively. $\sigma$ is the activation function (sigmoid here). Inspired by the general approach for link predictions, we aim to make the node features within the neighborhood more similar and distinguish them from the features of nodes outside the neighborhood by a loss function. So $M_s$ and $M_{sneg}$ are constructed and both used in calculating the objective function. 

To align the weights inside $M_s$ with $A^{(k)}$. A matrix $Idx \in \mathcal{R}^{2 \times N_k}$ containing edge indices of $A^{(k)}$ is utilized to transfer the shape of $M_s$. $\hat{M}_s \in \mathcal{R}^{N \times N}$ is the transferred mask of structure, $s_{ij} \in \hat{M}_s $ is computed as:
\begin{equation}
  s_{ij}=\begin{cases}
  M_{sk} & \text{ if } \quad Idx_{1,k}=i \quad \text{and} \quad Idx_{2,k}=j\\
  0 & \text{ otherwise }
\end{cases}
\label{ms_hat}
\end{equation}

\subsection{Explanation with Masks}
After explainable training, a matrix $M_f \in \mathcal{R}^{N \times F}$ is obtained from a well-trained mask generator and contains important weights for node features. By applying $M_f$ to the feature matrix $X$, we derive an explanation matrix $E_{feat} = M_f \odot X$ for the nonzero features of the node to provide feature explanations, where $\odot$ is Hadamard product represents the multiplication of the corresponding elements.

Similarly, the subgraph explanation $E_{sub}$, computed by $E_{sub} = \hat{M}_s \odot A^{(k)}$, provides important weights for the central nodes' k-hop neighbors. By covering these weights over the edges, SES provides explanations of the central node's neighbors in the subgraph.

\subsection{Construction of Positive-Negative Pairs}
To enhance the improvement of GNN's training, we extend supervisory information from the generated masks. Specifically, $M_s$ is used to construct positive and negative sample pairs for subsequent contrastive learning. 
This process involves a k-hop weight matrix, denoted as $\hat{A}^{(k)}$. For each node $v_i$ in $G$, we sort and sample $v_i$'s neighbors based on their weights in $\hat{A}^{(k)}$, forming the positive node set $S^p(v_i)$. In addition, we sample an equal number of nodes from $P_n(v_i)$ to construct the negative node set $S^n(v_i)$. The complete process is summarized in Algorithm \ref{posnegsample_pseudocode}, wherein \textbf{sorted} returns sorted neighbors of $v_i$ according to their weights in $\hat{A}^{(k)}$, $\textbf{len}$ returns the number of neighbors of $v_i$, and \textbf{random\_sample} randomly selects $num\_sample$ nodes from $P_n(v_i)$ to serve as negative samples.

\begin{algorithm}
\SetKwInput{KwData}{Input}
\SetKwInput{KwResult}{Output}
\KwData{Structure mask $\hat{M}_s$, k-hop adjacency $A^{(k)}$, negative node set $P_n$, sample ratio $r$, number of nodes $N$.}
\KwResult{The positive and negative node sets.}
Compute the k-hop adjacency weight matrix $\hat{A}^{(k)} = \hat{M}_s \cdot A^{(k)}$ \;
\For{$i$ to $N$}{
    neighs\_i = \textbf{sorted} ($\hat{A}^{(k)}_i$)\;
    num\_sample = r $\times$ \textbf{len} (neighs\_i)\;
    $S^p(i)$ = neighs\_i[0 : num\_sample]\;
    $S^n(i)$ = \textbf{random\_sample} ($P_n(v_i)$, num\_sample)\;
}
\Return positive and negative node sets $S^p$ and $S^n$\;
\caption{Construction of Positive-Negative Pairs.}
\label{posnegsample_pseudocode}
\end{algorithm}

\subsection{Learning Objective}
\label{learning_objective}
\subsubsection{Explainable Training}
During the explainable training phase, we simultaneously train the graph encoder and the mask generator.

For accuracy optimization, we minimize the cross-entropy loss for the graph encoder during training. The cross-entropy loss $\mathcal{L}_{xent}$ is defined as follows:
\begin{equation}
    \mathcal{L}_{xent}=-\sum_{l \in Y_L} \sum_{c=1}^{C}Y_{lc} \ln{Z_{lc}},
\end{equation}
where $Y_L$ denotes the node labels, $C$ denotes the class number of dataset, $Z$ is the output of graph encoder.

To ensure that the significant neighbors provided by $M_s$ conform to the original structure of the neighborhood, we design a subgraph loss. The loss stacks $M_s$ and $M_{sneg}$ together and minimizes the distance from the corresponding stacked labels $Y_s$ and $Y_{sneg}$. The formula for the subgraph loss $\mathcal{L}_{sub}$ is as follows:
\begin{equation}
    \mathcal{L}_{sub} = \frac{1}{N_k} \sum_{i=1}^{N_k} \left | stk(M_s, M_{sneg}) - stk(Y_s, Y_{sneg}) \right |,
    \label{Lsub}
\end{equation}
where $Y_s$ and $Y_{sneg}$ are neighboring nodes' labels and negative sample labels, respectively.

For reliable interpretations, $\hat{M}_s$ and $M_f$  are applied to features and adjacency to facilitate the training of the mask generator. The new output of the graph encoder is $Z_m$. Cross-entropy loss for $Z_m$ denotes $\mathcal{L}^m_{xent}$ that optimizes the neighborhood weights and feature weights suitable for the graph encoder.
The formulations are computed as follows:
\begin{equation}
    \begin{aligned}
        Z_m =& GE(M_f  \odot X, \hat{M}_s \odot A^{(k)}),\\
        \mathcal{L}^m_{xent} =& -\sum_{l \in Y_L} \sum_{c=1}^{C}Y_{lc} \log{Z_{mc}},
    \end{aligned}
    \label{Lmxent}
\end{equation}
where $GE$ denotes the graph encoder refers to Eq. (\ref{ge}).

To sum up, for the explainable training, the loss is:
\begin{equation}
    \min\limits_{\theta_m, \theta_e} [\alpha(\mathcal{L}_{sub} + \mathcal{L}^m_{xent}) +(1 - \alpha) \mathcal{L}_{xent}],
    \label{exp_training}
\end{equation}
where $\alpha$ is the hyperparameter to balance the optimization weights of the graph encoder and mask generator. $\theta_m$ and $\theta_e$ represent the learnable parameters of the mask generator and graph encoder, respectively.

\subsubsection{Enhanced Predictive Learning}
In the enhanced predictive learning phase, the structure mask $\hat{M}_s$ and feature mask $M_f$ generated by the trained mask generator are applied to the adjacency matrix $A$ and the node features $X$, respectively. This process highlights the critical components of structure and features, enhancing the graph encoder's representation ability. The output of $GE$ is denoted as $\hat{Z}$ and is computed as follows:

\begin{equation}
    \hat{Z} = GE(M_f \odot X, \hat{M}_s \odot A),
    \label{z_hat}
\end{equation}

To learn essential neighboring features from positive and negative sample pairs to enhance the node representation, a triplet loss $\mathcal{L}_{triplet}$ is designed. For each node $v_i$, we consider it as an anchor node and obtain the features of the positive, negative, and anchor samples mapped by $\hat{Z}$:
\begin{equation}
        p_i = \mathop{stk}\limits_{j=1}^{\mathcal{N}_i} \hat{Z}_{S^p(v_i)_j},\quad
        n_i = \mathop{stk}\limits_{j=1}^{\mathcal{N}_i} \hat{Z}_{S^n(v_i)_j 
              },\quad
        a_i = \mathop{stk}\limits_{j=1}^{\mathcal{N}_i} \hat{Z}_i,\\
    \label{construct_pos_neg}
\end{equation}
where $\mathcal{N}_i$ denotes the set number of node $v_i$ in $S^p$ and $S^n$.

The $\mathcal{L}_{triplet}$ is computed as follows:
\begin{equation}
    \mathcal{L}_{triplet} = \frac{1}{N} \sum_{i=1}^{N} \{ \max(\left \| a_i - p_i \right \|_2  - \left \| a_i - n_i \right \|_2  + m, 0)  \},
    \label{triplet_loss}
\end{equation}
where $m$ denotes the margin parameter in the triplet loss.

For enhanced predictive learning, the learning objective of SES is:
\begin{equation}
    \min\limits_{\theta_e} [\beta \mathcal{L}_{triplet} + (1 - \beta) \mathcal{L}_{xent}],
    \label{sup_training}
\end{equation}
where $\beta$ is the hyperparameter.


\begin{algorithm}
\SetKwInput{KwData}{Input}
\SetKwInput{KwResult}{Output}
\KwData{A graph $G$, hyperparameters $\alpha$ and $\beta$.}
\KwResult{The feature explanation $E_{feat}$ and substructure explanation $E_{sub}$. The nodes' representation $\hat{Z}$.}
Initialize the parameters $\theta_e$, $\theta_m$ of graph encoder and 
mask generator using Xavier initialization~\cite{xavier2010}\;
\While{epoch \textbf{in} explainable training}{
    Obtain the node representation $Z$ by the graph encoder (Eq. (\ref{ge}))\;
    Calculate the feature and structure mask $M_f$, $M_s$, and transferred structure mask $\hat{M}_s$ by the mask generator (Eqs. (\ref{mf}, \ref{ms}, \ref{ms_hat}))\;
    Update $\theta_e$, $\theta_m$ by the loss in explainable training (Eq. (\ref{Lmxent}, \ref{exp_training}))\;
}
Construct positive-negative node set $S^p$, $S^n$\;
\While{epoch \textbf{in} enhanced predictive learning}{
    Obtain the node representation $\hat{Z}$ by the graph encoder (Eq. (\ref{z_hat}))\;
    Calculate the positive-negative features by $\hat{Z}$ (Eq. (\ref{construct_pos_neg}))\;
    Compute the triplet loss (Eq. (\ref{triplet_loss}))\;
    Update $\theta_e$ by the loss in enhanced predictive learning (Eq. (\ref{sup_training}))\;
}
Return the output $\hat{Z}$, feature explanation $E_{feat}$ by $M_f$, substructure explanation $E_{sub}$ by $\hat{M}_s$\;
\caption{Framework of SES.}
\label{SES_pseudocode}
\end{algorithm}

\subsection{Time Complexity}

During the training process, the time consumption of the explainable training phase in SES contains the following main components: the backbone GNN, the computation of the mask generator, and Algorithm \ref{posnegsample_pseudocode}. In the case of backbone GCN~\cite{gcn2017}, the model complexity is $O(|E| \times F \times F_{hid})$. The Algorithm \ref{posnegsample_pseudocode} has a time complexity of $O(Nlog(N))$. For the feature mask $M_f$ in the mask generator, its time complexity is $O(F \times F_{hid})$. For the structure mask $M_s$ and the negative mask $M_{sneg}$ in the mask generator both have time complexities of $O(|V| \times N_k \times F_{hid})$. 
Similarly, the time complexity for the enhanced predictive learning phase is $O(|E| \times F \times F_{hid}) + O(F_{hid} \times \sum_{i=1}^{|V|}\mathcal{N}_i^2)$. Assuming sparsity in the graphs, we can simplify the complexities by considering $|E| = |V| \times \bar{K}_1/2 $, $N_k=|V| \times \bar{K}_2/2$, where $\bar{K}_1$ and  $\bar{K}_2$ is the average degree of nodes in $V$ and $A^{(k)}$ respectively. After discarding lower-order terms, the final time complexity for SES is $O(|V|^2\times \bar{K}_1 \times \bar{K}_2 \times F)$.

The time complexity of SEGNN~\cite{segnn2021} can be expressed as $O(F \times \sum_{v_t \in \mathcal{V}_t}\sum_{v_l \in \mathcal{S}_t}|E_{t}| \cdot|E_{l}|)$, where $\mathcal{V}_t$ is the set of target nodes, $\mathcal{S}_t$ denotes the sampled positive and negative nodes, and $E_l$ are the edges of the union set of $\mathcal{V}_t$ and $\mathcal{S}_t$. The number of nodes in the union set are comparable to $|V|$. The time complexity of SEGNN is $O(|V|^3 \times \bar{K}_3 \times \bar{K}_4 \times F)$, where $\bar{K}_3$ is the average degree of nodes of the union set, and $\bar{K}_4$ is the average number of the sample nodes. Neglecting the constants, the time complexity of SEGNN is higher than SES.

\section{Experimental Results}

\subsection{Datasets Description}
\subsubsection{Real-World Datasets}
Three classic datasets are considered: (1) \textbf{Cora}~\cite{plaintoid2016} is a citation network 
 with 2,708 nodes and 10,556 edges. Nodes represent documents, and edges represent citation links. Each paper is represented by a 1433-dimensional word vector and labeled into one of seven machine-learning topics.
(2) \textbf{CiteSeer}~\cite{plaintoid2016} is also a citation network with 3,327 nodes and 9,104 edges; it has six classes and the data structure is the same as Cora.
(3) \textbf{PolBlogs}~\cite{polblogs2005} is a graph with 1,490 nodes and 19,025 edges. A node represents political blogs, and edges represent links between blogs. Each node has a label that indicates its political inclination: liberal or conservative.
(4) \textbf{CS}: The Coauthor CS network from \cite{2018pitfall} with 18,333 nodes and 163,788 edges. Nodes represent authors that are connected by an edge if they co-authored a paper.

\subsubsection{Synthetic Datasets}
Following previous works~\cite{gnnexplainer2019,pgexplainer2020} to construct four synthetic datasets to validate the performance of GNNs on explainability tasks: (1) \textbf{BAShapes} contains a Barabasi-Albert graph with 300 nodes and a set of 80 “house”-structured graphs connected to it. The nodes are classified into four categories based on their structural roles. (2) \textbf{BACommunity} dataset is a union of BAShapes with two “house”-structure graphs. Nodes have normally distributed features and are assigned to one of eight classes based on their structural roles and community membership. (3) \textbf{Tree-Grid} contains nodes of the tree structure and grid motifs. The categories of nodes are the grid motif or tree structure they belong to. (4) \textbf{Tree-Cycle} is similar to the “Tree-Grid”, consisting of the tree structure and cycle motifs. 

\subsection{Baselines}
We consider the following strong baselines and verify whether the SES can improve the performance of backbone GNNs and achieve SOTA performance on the node classification tasks. We only report results of SES with GCN and GAT as backbone GNNs following ProtGNN~\cite{protgnn2022} and report the best performance of ProtGNN by GCN and GAT as backbone GNNs. PxGNN~\cite{pxgnn2022} has neither been published nor has public codes, making it unsuitable as a baseline.

\begin{itemize}
    \item GCN~\cite{gcn2017} is a typical graph neural network that integrates nodes' information from their neighbors.
    \item GAT~\cite{gat2018} fuses the attention mechanism based on GCN and enhances the ability of node representation.
    \item UniMP~\cite{transformconv2021} is a novel unified message-passing model incorporating feature and label propagation at training and inference time.
    \item SEGNN~\cite{segnn2021} can find K-nearest labeled nodes for each unlabeled node to give explainable node classification.
    \item FusedGAT~\cite{fusedgat2022} is an optimized version of GAT that fuses message-passing computation for accelerated execution and lower memory footprint.
    \item ProtGNN~\cite{protgnn2022} combines prototype learning with GNNs and provides a new perspective on the explanations of GNNs. 
    \item ASDGN~\cite{anti-symmetric2023} is a framework for stable and non-dissipative DGN design conceived through the lens of ordinary differential equations.
\end{itemize}

For the explainability tasks of GNNs, we consider the following strong baselines. Note that the SEGNN and ProtGNN are unsuitable for the feature explanation tasks.
\begin{itemize}
    \item GRAD~\cite{gnnexplainer2019} is a gradient-based method that computes the gradient of the GNN’s loss function for the adjacency matrix and the associated node features.
    \item ATT~\cite{gat2018} is a graph attention network that is used to provide explanations.
    \item GNNExplainer~\cite{gnnexplainer2019} is a GNN explanation method that maximizes the mutual information between a GNN’s prediction and distribution of possible subgraph structures to provide consistent and concise explanations for an entire class of instances.
    \item PGExplainer~\cite{pgexplainer2020} employs deep neural networks to parameterize the interpretation generation process, which makes it a natural way to explain multiple instances collectively.
    \item PGMExplainer~\cite{pgmexplainer2020} utilizes a generation probability model for graph data. This approach enables the model to learn concise underlying structures from observed graph data.
    \item GraphLIME~\cite{graphlime2022} is a model-agnostic, local, and nonlinear explanation method for GNN for node classification tasks motivated from LIME~\cite{lime2016}, which uses Hilbert-Schmit Independence Criterion (HSIC) Lasso, a nonlinear interpretable model. 
\end{itemize}

\subsection{Experimental Settings}
For the prediction task on node classification, we randomly divide the datasets as 60\% training set, 20\% validation set, and 20\% test set~\cite{gnnbaselines2022}. For the explanation task, the synthetic datasets are divided into an 80\% training set, 10\% validation set, and 10\% test set corresponding to the settings of~\cite{gnnexplainer2019}. For SES and baselines, the learning rate with the Adam optimizer is set to 0.003. The hidden layer size is set to 128. The sample ratio in Algorithm~\ref{posnegsample_pseudocode} is set to 0.8 and the margin $m$ in Eq. (\ref{triplet_loss}) is set to 1.0.

\subsection{Node Classification}

\begin{table*}[]
    \caption{Prediction Accuracy (\%) on Node Classification.}
    \renewcommand\arraystretch{1.3}
    \setlength\tabcolsep{5pt}
    \centering
    \scriptsize
    \begin{tabular}{c|ccccc|cccc|c}
    \toprule
        Methods & GCN & GAT & UniMP & FusedGAT & ASDGN &  SEGNN & ProtGNN & SES (GCN) &SES (GAT) & Imp.\\ \hline
        Cora & 86.83$\pm$2.98 & 86.81$\pm$1.36 & \underline{88.18$\pm$1.52} & 80.26$\pm$1.47 & 83.28$\pm$0.82 & 84.35$\pm$0.33 & 81.98$\pm$2.13 & \textbf{90.64$\pm$0.65} & 90.39$\pm$0.56 & 2.46 \\
        CiteSeer & 75.50$\pm$0.50 & 72.22$\pm$0.85 & 75.33$\pm$1.34 & 74.22$\pm$1.16 & 75.20$\pm$0.37 & \underline{76.10$\pm$0.74} & 73.42$\pm$1.68 & 78.51$\pm$0.86 & \textbf{78.69$\pm$0.84} &2.59  \\
        PolBlogs & 93.86$\pm$5.28 & 94.72$\pm$1.15 & \underline{95.45$\pm$0.91} & 94.63$\pm$0.99 & 80.45$\pm$1.89 & -- & 88.77$\pm$10.98 &  \textbf{97.90$\pm$0.55} & 97.86$\pm$ 0.47 & 2.45\\ 
        CS & 90.08$\pm$0.11 & 91.72$\pm$0.44 & 93.65$\pm$0.34 & 91.35$\pm$0.05 & \underline{93.70$\pm$0.12} & -- & 84.30$\pm$1.52 & \textbf{94.54$\pm$0.45} & 94.10$\pm$0.31 & 0.84 \\ \bottomrule
    \end{tabular}
    \label{nc}
      \vspace{-0.5cm}
\end{table*}

We evaluated the performance of SES and baselines on real-world datasets for node classification. The experimental results are summarized in TABLE~\ref{nc}. The second highest performance is highlighted with \underline{underline}. The Imp. represents improvements by SES compared to the best baseline. 

SES (GCN) and SES (GAT) denote SES using GCN and GAT as backbone GNNs, respectively. The time consumption of SES(GCN) on Cora, CiteSeer, PolBlogs, and CS are 10.5s, 12.3s, 13.1s, and 89.7s, respectively. The time consumption of SES(GAT) on Cora, CiteSeer, PolBlogs, and CS are 10.7s, 12.4s, 13.3s, and 92.2s, respectively. Notice that PolBlogs lacks node features. We assign a unit matrix as the node features, ensuring each node has an associated feature representation. SEGNN is not suitable for PolBlogs and CS. The SES framework achieved the SOTA performance and demonstrated superiority on all real-world datasets. SES outperformed the second-best method by a significant margin. SES has improvements of 2.46\%, 2.59\%, 2.45\%, and 0.84\% in absolute accuracy over the second-best method on Cora, CiteSeer, PolBlogs, and CS, respectively. While current self-explainable methods SEGNN and ProtGNN perform poorly in the tasks, not as well as the backbone GNNs in most cases. 

\subsection{Explanation Qualification}

We performed experiments on widely applied synthetic datasets to compare SES with other GNN explanation methods. Following the experimental settings in GNNExplainer~\cite{gnnexplainer2019}, we employed ground-truth explanations available for the synthetic datasets to quantify the accuracy of different explanation methods. The AUC scores for explanation tasks are summarized in TABLE~\ref{exp}.

SES performs superior on the BAShape and Tree-grid datasets over all other methods, demonstrating significantly enhanced interpretability. SES showcases the least relative improvement of 2.5\% on the BAShape dataset and 3.0\% on the Tree-grid dataset. SES also performs outstandingly on the Tree-Cycle dataset, achieving an AUC score close to 100\%. On the BACommunity dataset, both SES and PGExplainer achieved comparable performance. SEGNN performs well on the BAShapes while performing much less on the other three datasets. SES showcases its effectiveness in providing accurate and reliable explanations of synthetic datasets.

\begin{table}[!ht]
    \caption{Explanation accuracy (\%)  on Synthetic datasets.}
    \setlength\tabcolsep{5pt}
    \centering
    \begin{tabular}{c|cccc}
    \toprule
        Dataset & BAShapes & BACommunity & Tree-Cycle & Tree-Grid \\ \midrule
        GRAD & 88.2 & 75.0 & 90.5 & 61.2 \\
        ATT & 81.5 & 73.9 & 82.4 & 66.7 \\
        GNNExplainer & 92.5 & 83.6 & 94.8 & 87.5 \\
        PGExplainer & 96.3 & \textbf{94.5}& \underline{98.7} & \underline{90.7} \\
        PGMExplainer & 96.5 & \underline{92.6} & 96.8 & 89.2 \\
        SEGNN & \underline{97.3} & 77.2 & 62.3 & 50.5\\\midrule
        SES & \textbf{99.8} (2.5$\uparrow$) & \textbf{94.5} & \textbf{99.4} (0.7$\uparrow$) & \textbf{93.7} (3.0$\uparrow$) \\ \bottomrule
    \end{tabular}
    \label{exp}
\end{table}

To evaluate the feature explanation quality on real-world datasets, we consider SES and two classical methods, GNNExplainer~\cite{gnnexplainer2019} and GraphLIME~\cite{graphlime2022}, which are suitable for the node feature explanation. However, PGExplainer and PGMExplainer, target for edge and node explanations, respectively, lack the capability to generate the weight set for the features, rendering them unsuitable for the present task. Fidelity+~\cite{fidelity+2019}, a widely used metric, is employed. Fidelity+ measures the dissimilarity in accuracy or predicted probability between the original predictions and the predictions obtained after masking out crucial input features. Mathematically, it is expressed as follows:
\begin{equation}
    \mathrm{Fidelity+}^{acc}=\frac{1}{N}\sum_{i=1}^{N}\left(1\left(\hat{y}_{i}=y_{i}\right)-1\left(\hat{y}_{i}^{1-m_{i}}=y_{i}\right)\right)
\end{equation}
where $y_i$ represents the original prediction, and $1-m_i$ denotes the complementary mask that eliminates the important input features. The indicator function $1 \left( \hat{y}_i=y_i \right )$ evaluates to $1$ when $y_i$ and $\hat{y}_i$ are equal, and $0$ otherwise.

Due to the sparsity of the citation network, we remove the top-5 important features of each node based on the weights assigned by explainers. We conduct experiments by 2-layer GCN and GAT with identical parameters. The Fidelity+ scores are presented in TABLE~\ref{fidelity}. The importance weights assigned by GraphLIME do not significantly influence node classification's performance, as the accuracy only experiences a minor decline after removing the important features. 

SES achieves the highest Fidelity+ scores on Cora, CiteSeer, Polblogs, and CS datasets. The Fidelity+ of SES (GCN) is about four times of GNNExplainer (GCN) on CiteSeer. When GAT is used as the backbone GNN, SES (GAT) and GNNExplainer (GAT) represent comparable performance on the PolBlogs. The gaps in performance are also reduced on Cora and CiteSeer. 

The post-hoc methods GNNExplainer and GraphLIME utilize a separate training process to generate explanations, which may result in disjointed explanations. The feature and structure masks in SES directly work on the feature and adjacency matrices, respectively, and they are co-trained with the backbone GNN by Eq. (\ref {Lmxent}). It ensures feature and structure masks are consistent with the backbone GNN's aggregation and decision process. To validate the reason, we removed $\mathcal{L}^m_{xent}$ from Eq. (\ref {Lmxent}), which is denoted as $-\{\mathcal{L}^m_{xent}\}$. The results demonstrate a significant performance decay for SES after eliminating the impact of the consistency, which indicates the effectiveness of the proposed mask generator.

\begin{table}[]
    \caption{Fidelity+ (\%) of models on real-world datasets.}
    \centering
    \begin{tabular}{c|cccc}
    \toprule
        Dataset & Cora & CiteSeer & PolBlogs & CS\\ \midrule
        GNNExplainer (GCN) & 8.3 & 4.3 & 40.5 & 0.17\\
        GraphLIME (GCN) & 1.6 & 1.7 & 2.0 & 0.09\\
        SES (GCN)$-\{\mathcal{L}^m_{xent}\}$ & 5.27 & 1.79 & 48.53 & 0.6\\
        SES (GCN) & \textbf{14.7} & \textbf{16.1} & \textbf{49.3} & \textbf{2.77}\\ \hline
        Imp.  & 6.4 & 11.8 & 8.8 & 2.17\\ \midrule
        GNNExplainer (GAT) & 15.4 & 9.4 & \textbf{44.8} & 0.15\\
        GraphLIME (GAT) & 1.2 & 1.0 & 2.8 & 0.12 \\
        SES (GAT)$-\{\mathcal{L}^m_{xent}\}$ & 1.30 & 2.17 & 39.13 & 0.3\\
        SES (GAT) & \textbf{17.2} & \textbf{11.0} & 44.6 & \textbf{2.96}\\ \hline
        Imp.   & 1.8 & 1.6 & -0.2 & 2.66\\ \bottomrule
    \end{tabular}
    \label{fidelity}
      \vspace{-0.5cm}
\end{table}

\subsection{Time Consumption}

We evaluate and compare the time consumption post-hoc explainers and self-explainable GNNs to generate explanations on Cora. Traditionally speaking in machine learning, inference time is the duration time from the input to a well-trained model to its producing outputs, which are milliseconds and is not particularly meaningful for comparing GNNs. we define inference time for generating explanations as the period required from a well-trained backbone GNN to produce explanations for all nodes following the way of PGExplainer \cite{pgexplainer2020}. For GNNExplainer and GraphLIME, despite backbone GNNs have been trained, they necessitate re-training for each individual node \cite{gnnexplainer2019, graphlime2022}, so the inference time counts on the re-training duration. For SES and SEGNN, the inference time incorporates the training time since they train backbone GNN models while giving explanations for the same process. This allows for a fair comparison of the time consumption between post-hoc methods and self-explainable GNNs. 
However, ProtGNN cannot construct explainable subgraphs for node classification tasks, making it inapplicable for this specific task. The explainable training of SES spans 300 epochs with an additional 15 epochs dedicated to enhanced predictive learning. The training epochs for other explainers are aligned with SES. The experiments were on an NVIDIA RTX 3090 GPU, with GCN as the backbone GNN, and are presented in TABLE~\ref{expl_time_consumption}.

\begin{table}[!ht]
    \caption{Inference time of generating explanations for all nodes on the Cora dataset.}
    \setlength\tabcolsep{5pt}
    \centering
    \begin{tabular}{ccccc}
    \toprule
         GNNExplainer & GraphLIME &PGExplainer&SEGNN & SES ($et$) \\\midrule
          9 min 50s & 4 min 24s & 1 min 13s & 1 min 32s & 4.3s\\\bottomrule
    \end{tabular}
    \label{expl_time_consumption}
\end{table}

TABLE~\ref{expl_time_consumption} showcases that SES outpaces SEGNN and other post-hoc explainers in terms of speed. The mask of SES is co-trained with the graph encoder and subsequently provides explanations once the explainable training (SES ($et$)) is completed, clocking in at a swift 4.3 seconds. The enhanced predictive learning phase of SES (SES $(epl)$) that involves conducting contrastive learning for each node takes 6.5 seconds. The total training time for the whole SES is 10.8 seconds. Note that SES $(epl)$ does not affect the explainability of SES but refines its prediction accuracy. The training and inference times for SES (GCN) across real-world datasets, as detailed in Table \ref{traintime}, exhibit an increase corresponding to the growing graph sizes and densities.

\begin{table}[]

    \caption{Training and inference time of SES(GCN).}
    \centering
    \begin{tabular}{c|cccc}
    \toprule
        Dataset & Cora & CiteSeer & PolBlogs & CS\\ \midrule
       
        Inference time & 4.3s &	4.4s &	9.1s &	34.0s\\ 
    
        Training time & 10.8s &	12.3s &	13.1s &	89.7s \\ \bottomrule
    \end{tabular}
    \label{traintime}
    \vspace{-0.3cm}
\end{table}

The time complexity of Algorithm \ref{posnegsample_pseudocode} is primarily dictated by the sorting and random sampling operations that have $O(NlogN)$ and $O(N)$ complexity, respectively, which only contributes a minor fraction of overall time consumption in SES. We calculate the time consumption of the Algorithm \ref{posnegsample_pseudocode} by synthesizing a sparse graph with a fixed number of nodes and twice as many edges, which is reported in TABLE \ref{pn_time_consumption}.

\begin{table}[]
    \caption{Time consumption of constructing positive-negative node pairs.}
    \centering
    \begin{tabular}{c|cccccc}
    \toprule
        Nodes & 0.1k  & 1k & 10k & 50k & 70k \\ \midrule
        Time consumption & 0.005s & 0.045s & 2.11s & 28.92s & 38.53s\\\bottomrule
    \end{tabular}
    \label{pn_time_consumption}
\end{table}

In the trade-off analysis, post-hoc methods exhibit competitive explanation accuracy compared to current self-explainable models but fall short or underperform in predicting node labels. Self-explainable methods SEGNN and SES offer lower inference time in comparison to GNNExplainer and GraphLIME that necessitate retraining for each node's explanation generation. SES demonstrates that high prediction accuracy and explanation quality can be achieved simultaneously. With the application of appropriate techniques, time efficiency need not be compromised. But SES and SEGNN come with the trade-off of higher memory demands which will be optimized in future work. In contrast, GNNExplainer and GraphLIME require smaller memory by interpretating on individual instances.  

\subsection{Parameter Sensitivity}
\begin{figure}[h]
\vspace{-0.5cm}
  \centering
  \subfigure[Learning rate / k-hop SES (GCN)]{
    \label{gcn_lr_khop}
    \includegraphics[scale=0.35]{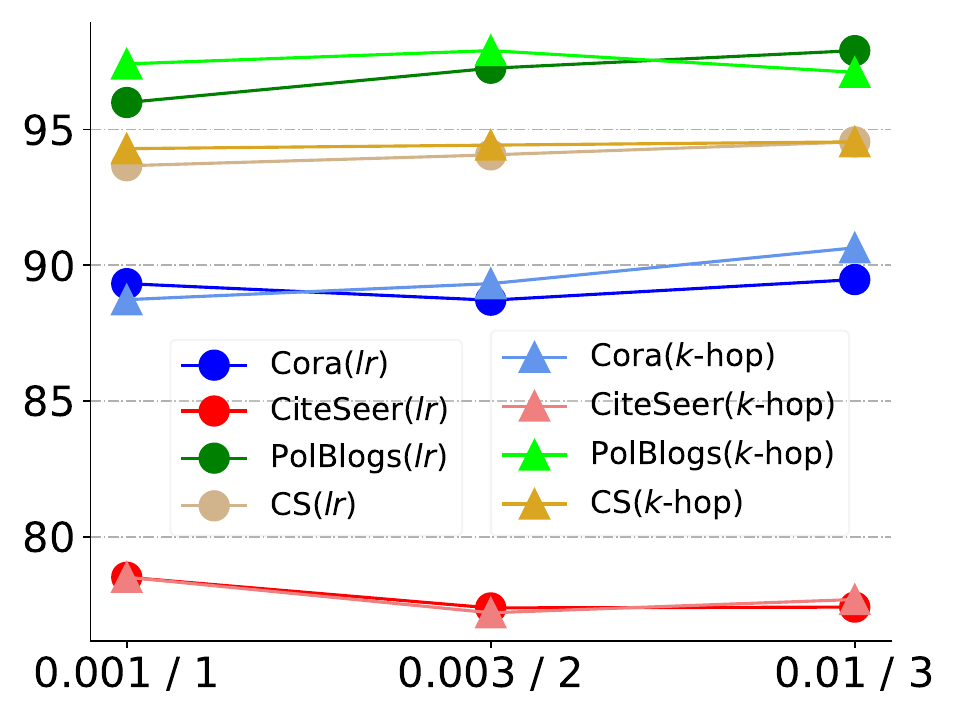}
  }
  \subfigure[Hyperparameter SES (GCN)]{
    \label{gcn_alpha_beta}
    \includegraphics[scale=0.35]{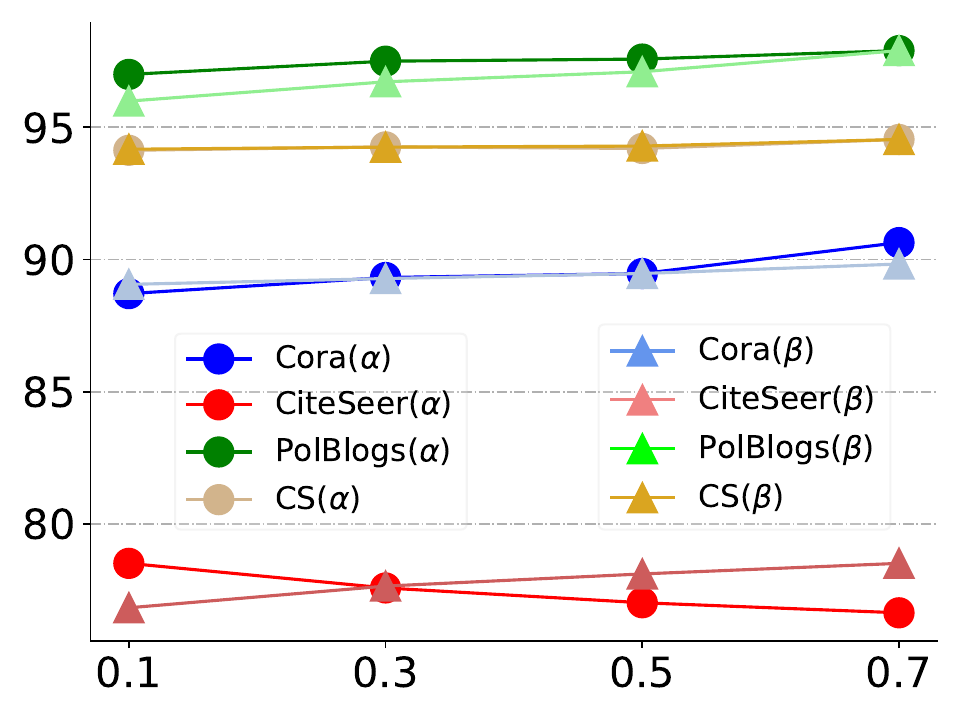}
  }
  \subfigure[Learning rate / k-hop SES (GAT)]{
    \label{gat_lr_khop}
    \includegraphics[scale=0.35]{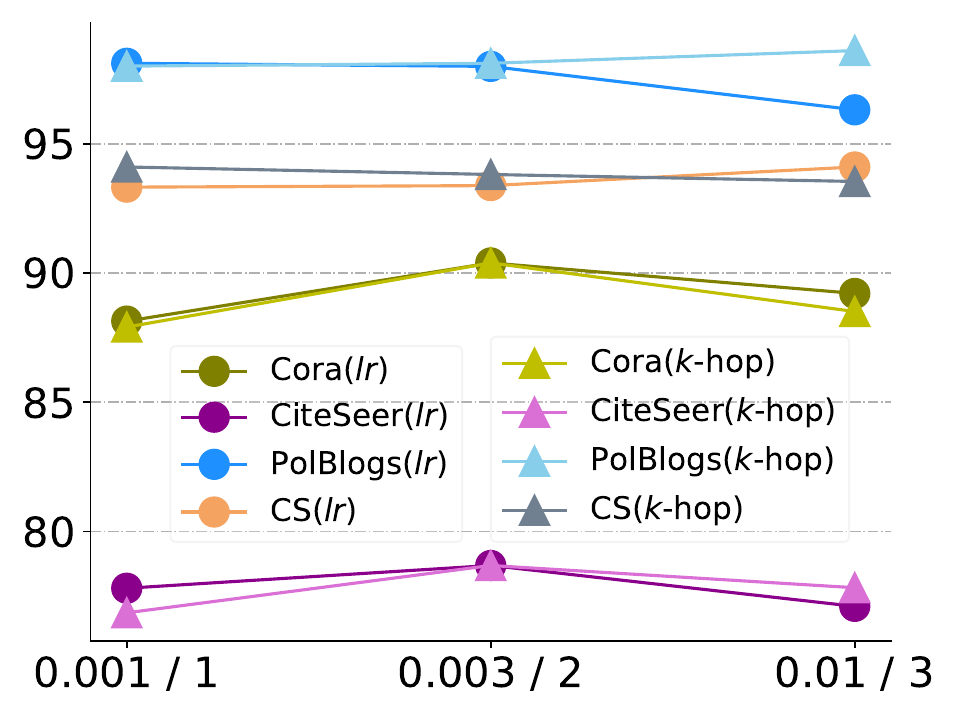}
  }
  \subfigure[Hyperparameter SES (GAT)]{
    \label{gat_alpha_beta}
    \includegraphics[scale=0.35]{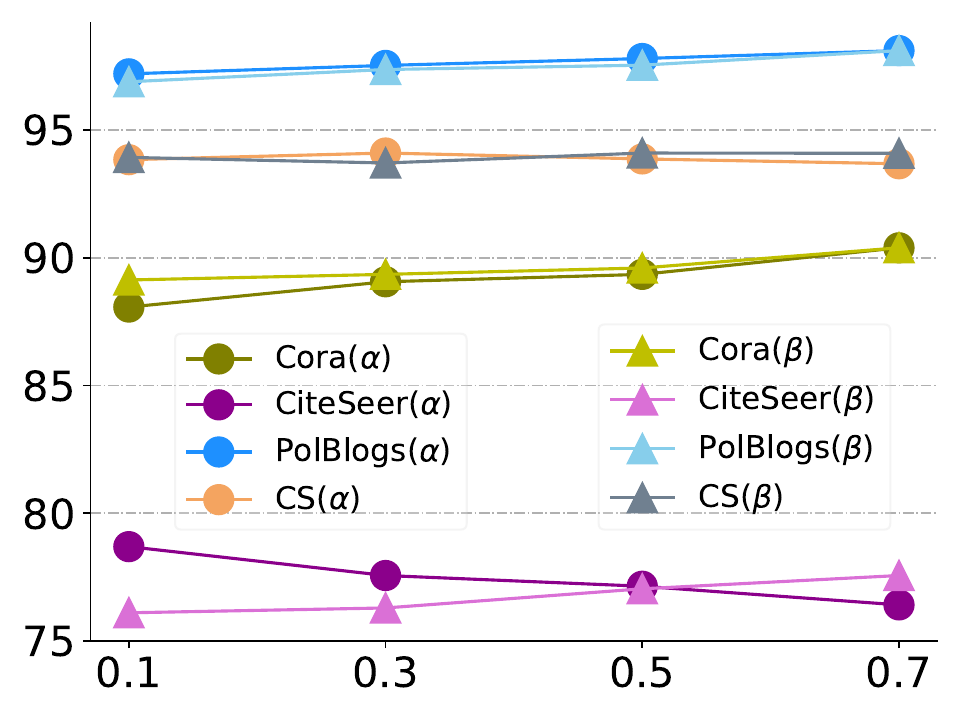}
  }
  \caption{Parameter sensitivities of SES.}
  \label{para_sen}
\end{figure}
The parameter sensitivity analyses are presented in Fig.~\ref{para_sen}, the learning rate ($lr$), $k$ (k-hop), and hyperparameters $\alpha$ and $\beta$ exhibit varying degrees of influence on SES's performance.

In Fig.~\ref{gcn_lr_khop}, SES (GCN)'s accuracy on CiteSeer improves as the learning rate and $k$ decrease. In the case of a larger neighborhood in Cora, SES (GCN) performs well owing to the wider range of papers cited from different fields.
Additionally, the effect on the PolBlogs increases with the learning rate. Fig.~\ref{gcn_alpha_beta} indicates that assigning higher weights to $\alpha$ and $\beta$ enhances the performance of SES (GCN) on Cora and PolBlogs. It suggests that mask training and self-supervised learning contribute to characterizing GNNs. However, a lower $\alpha$ weight is required for CiteSeer.

From Fig.~\ref{gat_lr_khop}, SES (GAT) consistently achieves the best performance on citation (Cora and CiteSeer) datasets when the learning rate is 0.003. On the PolBlogs, the value $k$ that leads to a larger neighborhood of a node and a small learning rate benefit the performance. Fig.~\ref{gat_alpha_beta} shows that the performance of the two hyperparameters exhibits a slow increase trend on Cora and PolBlogs. In the case of CiteSeer, SES (GAT) relies more on triplet loss to achieve optimal performance. However, the performance of SES is stable and notable in most cases. On the CS dataset, the performance of SES (GCN) and SES (GAT) is not sensitive to hyperparameters. Overall, we observe that higher learning rates and smaller neighborhood ranges lead to improved accuracy in SES.

\subsection{Visualization}

\begin{figure}[h]
  \centering
  \subfigure[SES (GCN)]{
    \label{gex_bs}
    \includegraphics[scale=0.45]{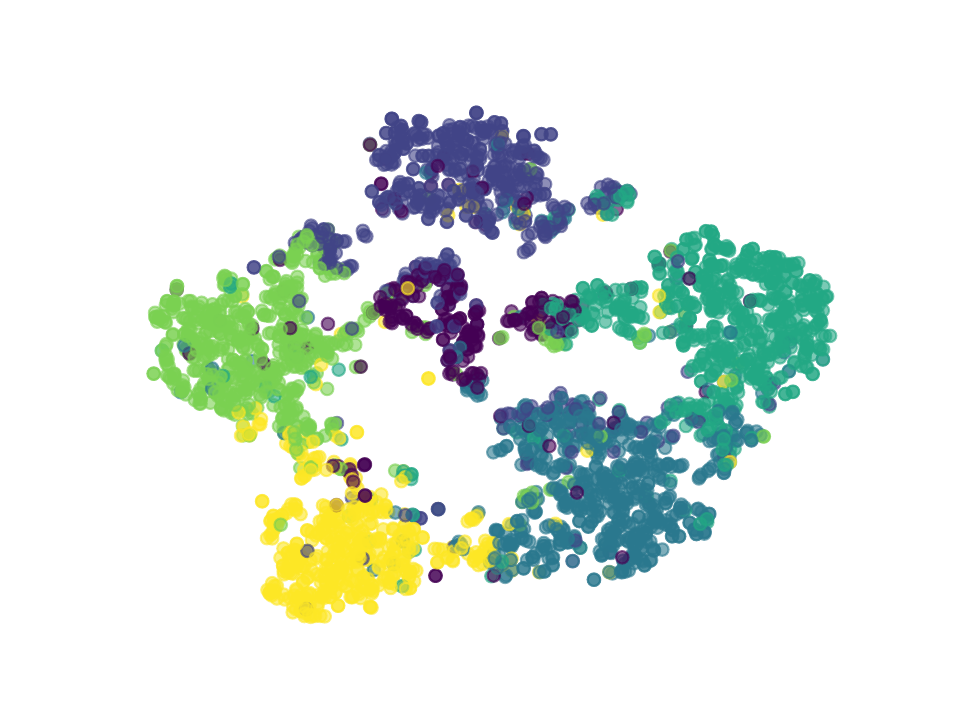}
  }
  \subfigure[SES (GAT)]{
    \label{gex_bc}
    \includegraphics[scale=0.45]{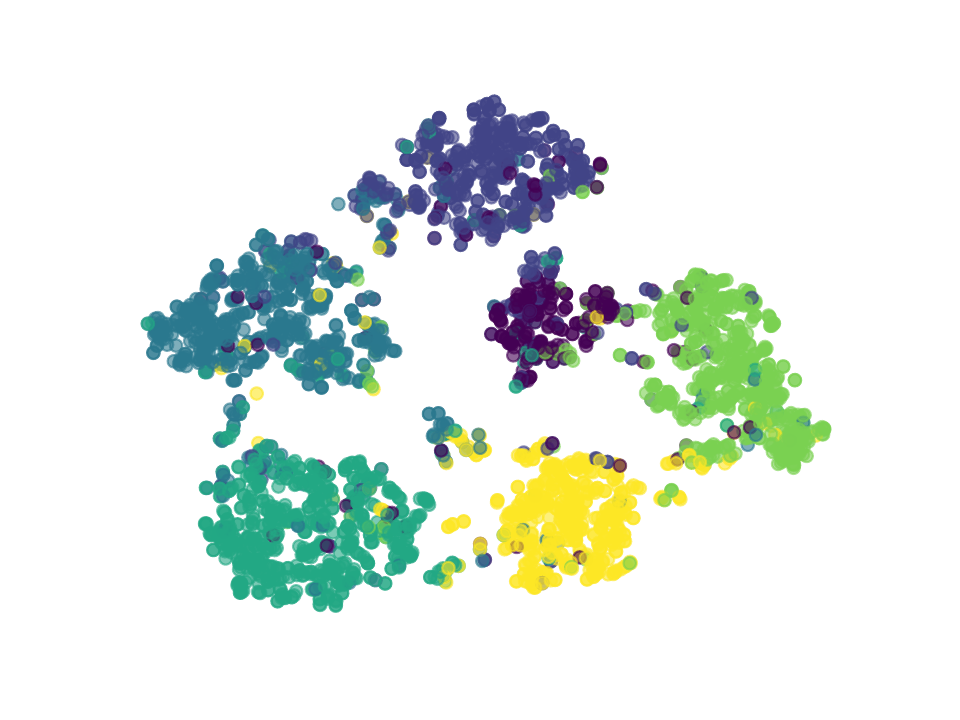}
  }
  \subfigure[SEGNN]{
    \label{gex_bs_2}
    \includegraphics[scale=0.4]{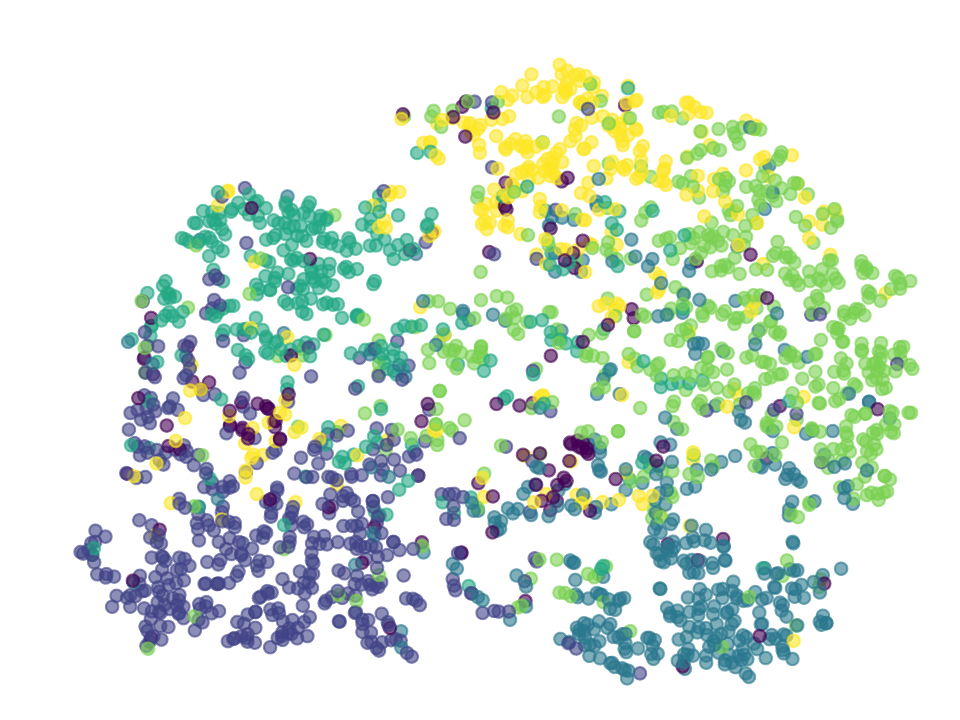}
  }
  \subfigure[ProtGNN]{
    \label{gex_bc_2}
    \includegraphics[scale=0.4]{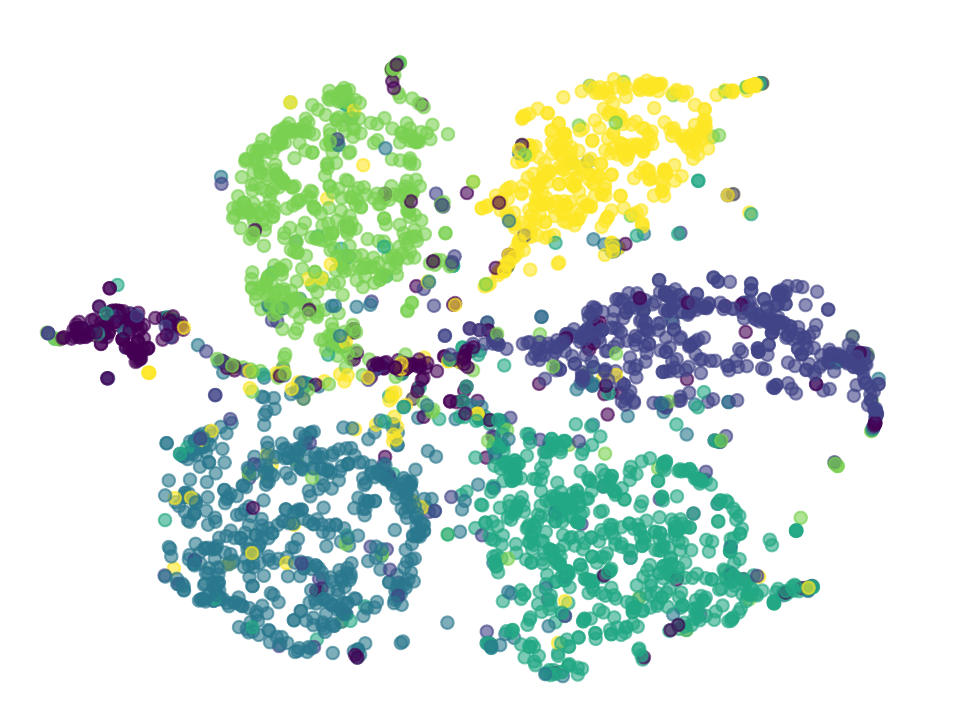}
  }
  \caption{Visualization of node representations after training on Citeceer.}
  \label{vis_nc}
    \vspace{-0.3cm}
\end{figure}

\begin{figure*}[h]
  \centering
  \includegraphics[scale=0.5]{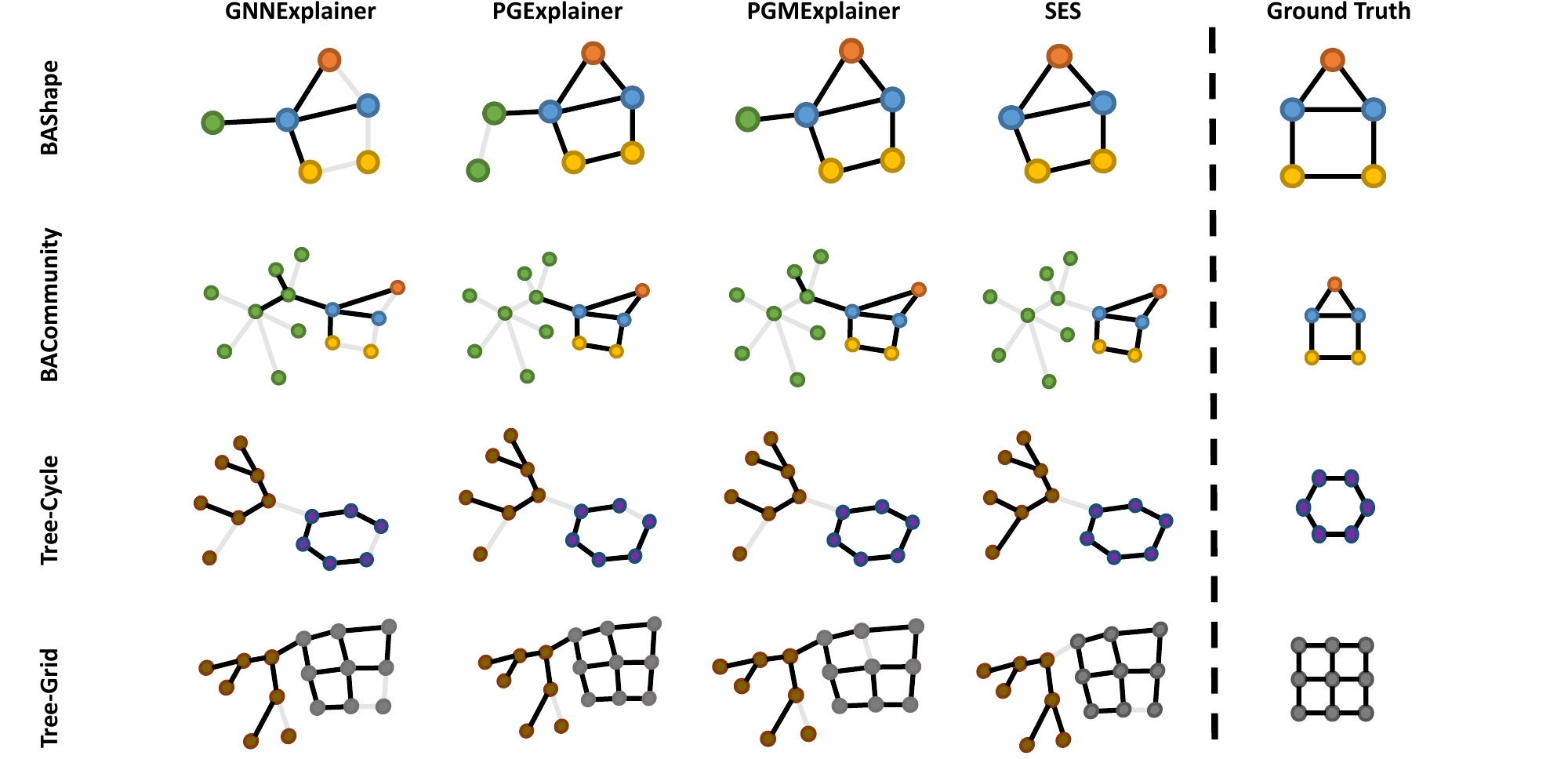}
  \caption{Visualizations of subgraph explanations on synthetic datasets.}
  \label{vis_subgraph}
    \vspace{-0.5cm}

\end{figure*}

\begin{table}[]
    \caption{Statistical metrics for visualization on CiteSeer.}
    \centering
    \begin{tabular}{c|cc}
    \toprule
        Metric & Silhouette~\cite{silhouettes1987} & Calinski harabasz~\cite{caha1974} \\ \midrule
        SES (GCN) & 0.316 & 1694.75\\
        SES (GAT) & \textbf{0.375} & \textbf{2131.56}\\\midrule
        SEGNN &  0.131 & 456.37\\ 
        ProtGNN &  0.277 & 1090.13\\\bottomrule
    \end{tabular}
    \label{vis_qua}
      \vspace{-0.5cm}
\end{table}

We visualized the learned node representations of GNNs to validate their representation ability on the CiteSeer dataset. Two self-explainable GNN models (SEGNN and ProtGNN) are applied. The output vectors of these GNNs are originally 128-dimensional and transformed into two dimensions by TSNE~\cite{tsne2008}. Fig.~\ref{vis_nc} depicts the visualizations of node representations. We observe that SES (GCN) and SES (GAT) provide more densely connected clusters than baselines. To qualify the clustering effectiveness of different methods, classical clustering evaluation indicators  Silhouette score~\cite{silhouettes1987} and Calinski Harabasz score~\cite{caha1974} are employed, with higher values indicating better outcomes. TABLE~\ref{vis_qua} presents the assessment of cluster effects following the visualizations of Fig.~\ref{vis_nc}. From TABLE~\ref{vis_qua}, SES exhibits tighter clusters among nodes belonging to the same classes, resulting in higher scores.

To comprehensively evaluate the effects of different explanation methods, we visualize the explanations of GNNExplainer, PGExplainer, PGMExplainer, and SES on four synthetic datasets \cite{gnnexplainer2019}. These visualizations highlight the important subgraph structures of the datasets. In the visualizations, the color of the edges in the corresponding graphs represents the importance weight, with darker edges indicating higher importance. The visualizations are presented in Fig.~\ref{vis_subgraph}. The explanations of SES on BAShape and BACommunity are evident that SES effectively identifies and matches important “house” structures in the datasets. For the Tree-Cycle and Tree-Grid datasets, SES also successfully recognizes the “tree”,  “circle”, and “grid” nodes on the Tree-Cycle and Tree-Grid datasets. As shown in the figures, the interpretation results from baselines include subgraphs with unrelated structures.

\subsection{Ablation Studies and Variants}

Ablation studies and variants are conducted to investigate the contributions of components in SES. To verify the functionality of the feature mask on GNN, we remove the $M_f$ on node features and denote it as $-\{M_f\}$. Similarly, a general adjacency matrix rather than $\hat{M}_s$ on adjacency put into the graph encoder during the enhanced predictive learning phase is denoted as $-\{\hat{M}_s\}$. To verify the guiding effect of cross-entropy on triplet loss, we remove $\mathcal{L}_{xent}$ and denote it as $-\{\mathcal{L}_{xent}\}$ in the enhanced predictive learning phase. To verify the effectiveness of enhanced predictive learning, the triplet loss is removed, denoted as $-\{Triplet\}$. To validate the importance of explainable training, we replace the mask generator with the post-hoc GNNs (GNNExplainer (GEX), PGExplainer (PGE)) and denote them as $+\{epl\}$. The performance of ablation studies is reported in TABLE~\ref{ablation}.
The performance of SES was significantly impacted after removing the contrast learning. It suggests that self-supervised learning is significant in the prediction performance of SES, and the interpretation results play a crucial role in providing feedback and enhancing the predictive learning of models. The cross-entropy also plays an important role in enhancing predictive learning since removing $\mathcal{L}_{xent}$ in the second phase in SES leads to a sharply decreased accuracy, especially for the CS dataset. 
The absence of $M_f$ and $\hat{M}_s$ both results in a decline in performance. This indicates that interpreting node features and structure importance plays significant roles in SES. Integrating self-supervised learning and the mask generator that considers structural and feature explanations leads to the best performance in SES. The results of $+\{epl\}$ combined with masks using the post-hoc model are worse than SES and most variants of SES. This indicates that the explanations provided by SES are more reliable and better suited for downstream constrastive learning.

\begin{figure*}[]
  \centering
  \includegraphics[scale=0.25]{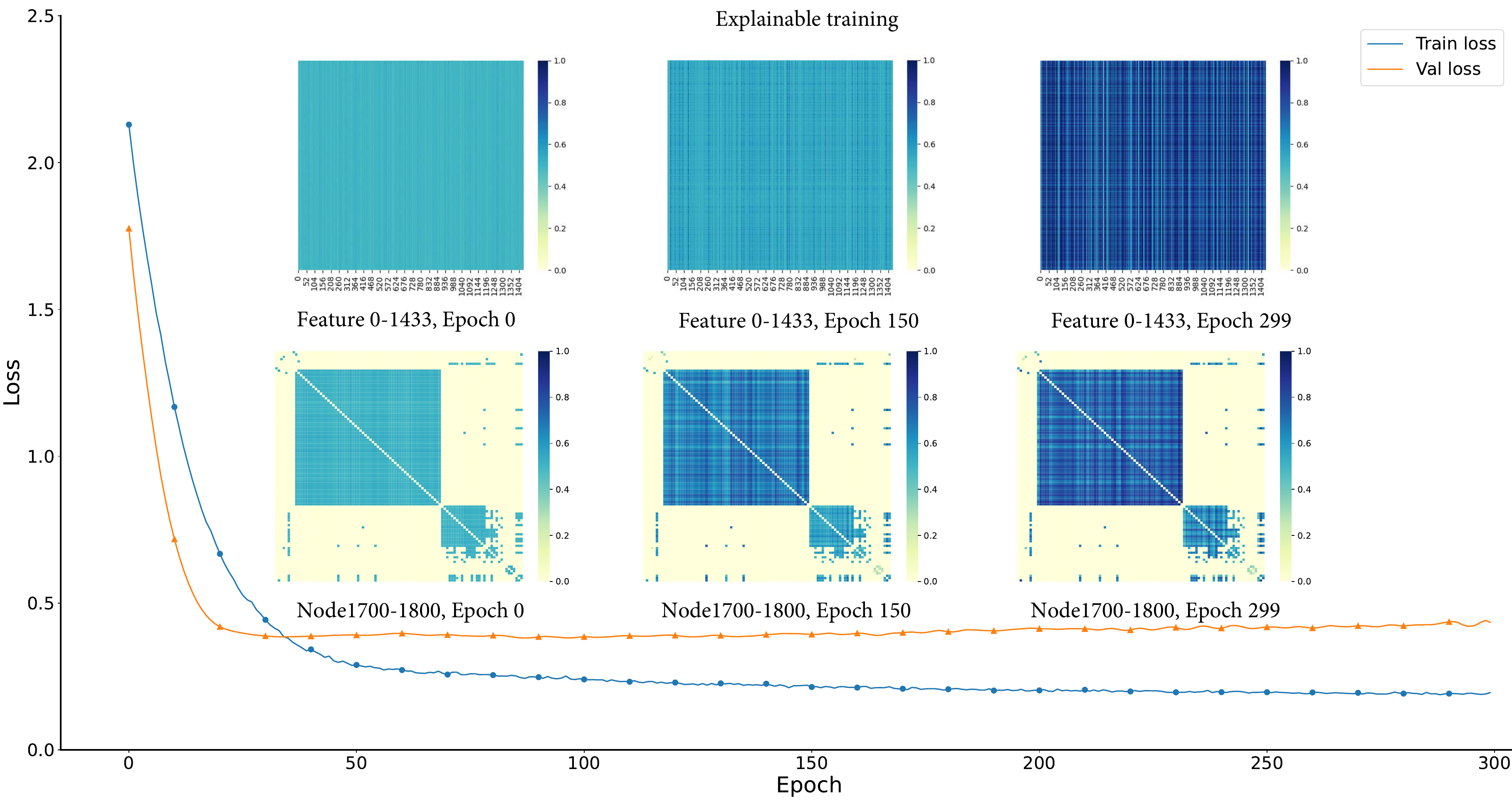}
  \caption{Optimization of the feature and structural masks during the explainable training on the Cora dataset.}
  \label{loss_curve}
    \vspace{-0.5cm}
\end{figure*}

\begin{table}[]
    \caption{Ablation studies of SES on real-world datasets.}
    \centering
    \begin{tabular}{l|cccc}
    \toprule
        Dataset & Cora & CiteSeer & PolBlogs & CS \\\midrule
        SES (GCN)$-\{M_f\}$ & 90.05 & 77.29 & 97.41 & 93.78 \\
        SES (GCN)$-\{\hat{M}_s\}$ & 89.31 & 78.05 & 96.90 & 94.24\\
        SES (GCN)$-\{\mathcal{L}_{xent}\}$ & 88.90 & 77.23 & 95.89 & 88.20\\
        SES (GCN)$-\{Triplet\}$ & 88.31 & 76.80 & 95.42 & 93.26\\\midrule
        GEX(GCN)$+\textit{\{epl\}}$  & 75.51 & 74.73 & 92.16 & 92.72\\
        PGE (GCN)$+\textit{\{epl\}}$ & 87.48 & 75.64 & 95.05 & 93.08\\\midrule
        SES (GCN) & \textbf{90.64} & \textbf{78.51} & \textbf{97.90} & \textbf{94.54}\\ \midrule
        SES (GAT)$-\{M_f\}$ & 89.56 & 77.29 & 96.98 & 92.86\\
        SES (GAT)$-\{\hat{M_s}\}$ & 88.29 & 77.41 & 95.14 & 92.91 \\
        SES (GAT)$-\{\mathcal{L}_{xent}\}$ & 88.43 & 77.93 & 96.24 & 88.28\\
        SES (GAT)$-\{Triplet\}$ & 87.81 & 76.76 & 94.81 & 91.60\\\midrule
        GEX(GAT)$+\textit{\{epl\}}$  & 83.61 & 71.69 & 95.63 & 91.86\\
        PGE (GAT)$+\textit{\{epl\}}$  & 87.66 & 72.60 & 95.06 & 92.45\\\midrule
        SES (GAT) & \textbf{90.39} & \textbf{78.69} & \textbf{97.86}& \textbf{94.10}\\ \bottomrule
    \end{tabular}
    \label{ablation}
      \vspace{-0.3cm}
\end{table}

\subsection{Case Studies}
\begin{figure}[!htp]
  \centering
  \subfigure[Cora, central node: 78]{
    \label{case_cora}
    \includegraphics[scale=0.45]{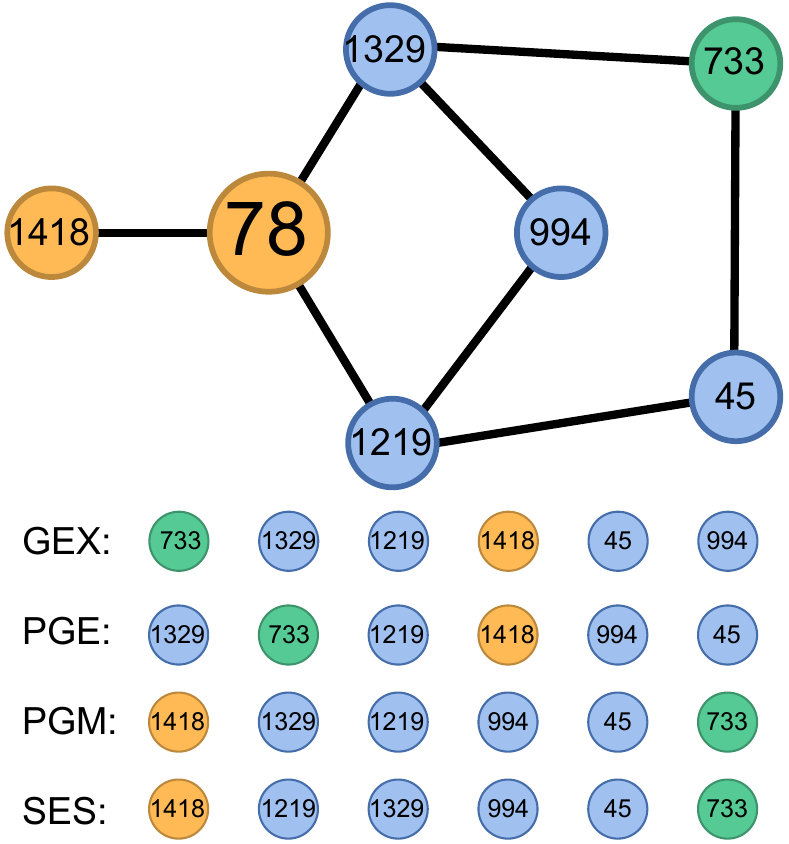}
  }
  \subfigure[CiteSeer, central node: 50]{
    \label{case_citeseer}
    \includegraphics[scale=0.45]{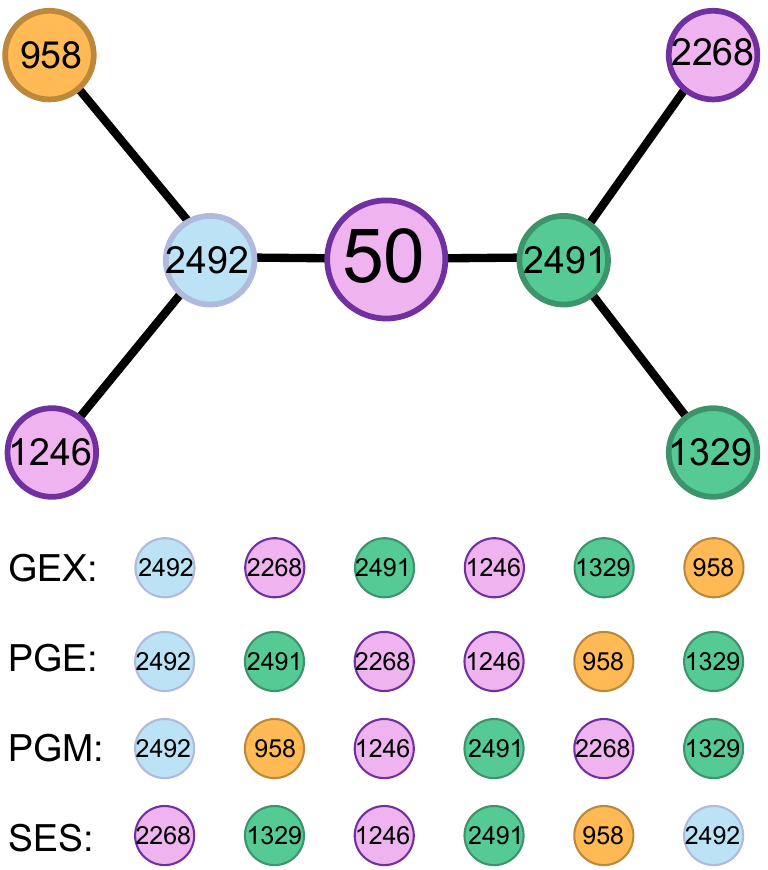}
  }
  \subfigure[PolBlogs, central node: 539]{
    \label{case_polblogs}
    \includegraphics[scale=0.45]{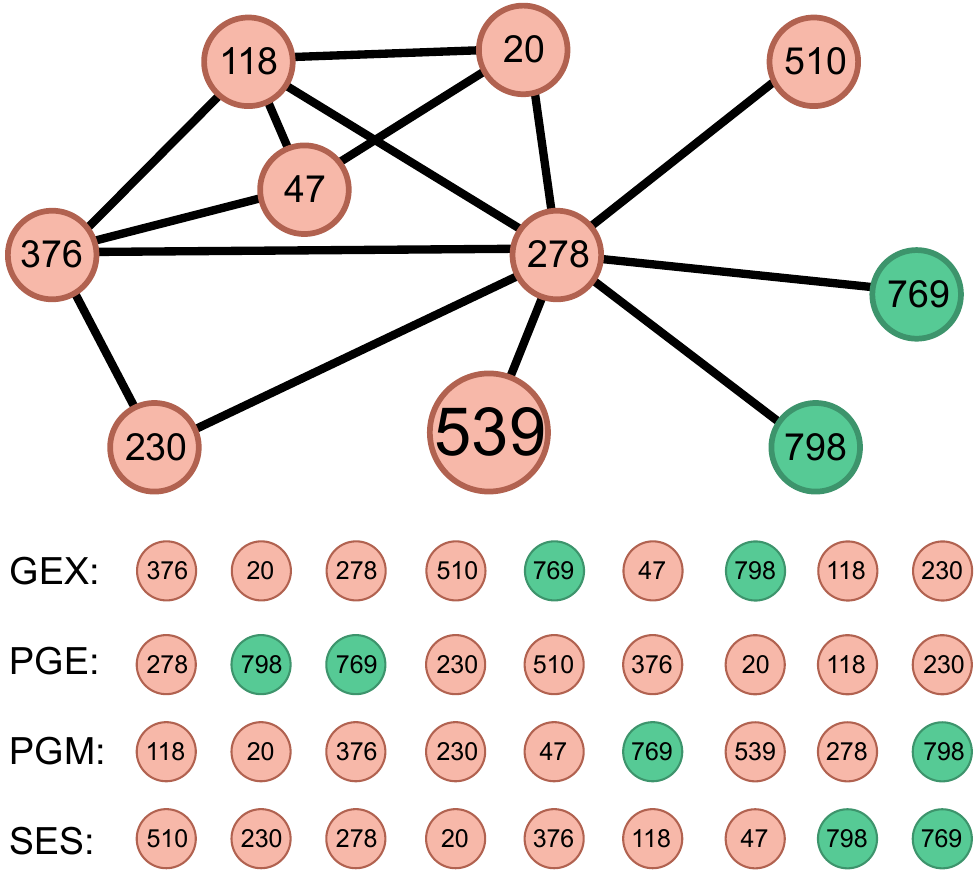}
  }
  \subfigure[CS, central node: 212]{
    \label{case_cs}
    \includegraphics[scale=0.45]{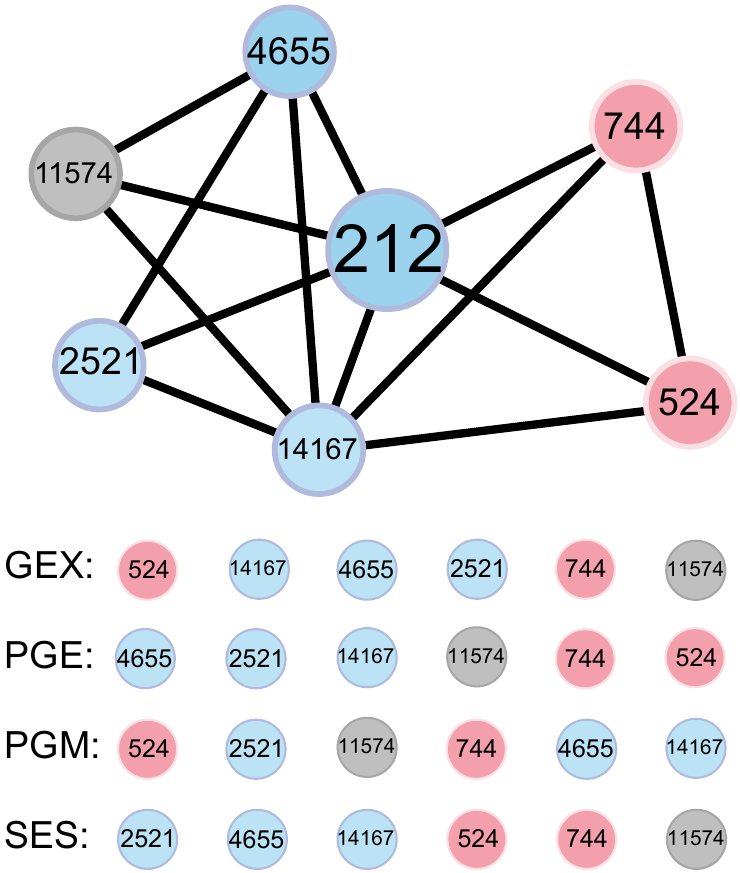}
  }
  \caption{Case study of subgraph explanations on real-world datasets. The rank of neighbors is based on the structure mask $\hat{M}_s$ of SES and edge masks of GNNExplainer (GEX), PGExplainer (PGE), and PGMExplainer (PGM). Different colored nodes have different labels.}
  \label{case_study}
\end{figure}

For the real-world datasets, we visualize the subgraphs by potential neighbor scores produced from the $\hat{M}_s$ of SES and edge masks of baselines to demonstrate their rationality in explanations. Especially, SES ranks the important neighbors with the weights obtained by $\hat{M}_s$, and the other baselines (GNNExplainer, PGExplainer, and PGMExplainer) rank important neighbors by the weights of edge masks for the central node in the neighborhood. Fig.~\ref{case_study} draws the 2-hop subgraphs of node 78 in Cora, node 50 in CiteSeer, node 539 in PolBlogs, and node 212 in CS and lists the ranked node sequence of models. 

In Fig.~\ref{case_cora}, SES and PGMExplainer put node 1418 as the most important neighbor of central node 78 since the nodes belong to the same class. In contrast, node 1418 is ranked with much lower importance in GNNExplainer and PGExplainer. In Fig.~\ref{case_citeseer}, the neighborhood of the central node contains multiple classes, which challenges ranking the important neighbors. SES ranks the nodes with the same class at the top of the sequence, while baseline methods fail to achieve it. In Fig.~\ref{case_polblogs}, SES assigns the lowest importance to the political pages of the green camp and produces more weight to nodes with the same class, while PGExplainer considers more important pages from the opposing party. In Fig.~\ref{case_cs}, SES and PGExplainer ranked nodes in the same classes as 212 in the top position, while GEX and PGM rated node 524 as more important. The cases demonstrated that SES provides more reasonable explanations than GNNExplainer, PGExplainer, and PGMExplainer.

\subsection{Mask Optimization}
We analyze the training and validation loss curves presented in Figure \ref{loss_curve}, as well as the evolution of the feature and structure masks during the training phase.  The feature mask is presented for all dimensions of all nodes at epochs 0, 150, and 299. Moreover, we draw the structural mask of the 2-hop for nodes ranging from 1700 to 1800.
In Figure \ref{loss_curve}, the initial state of the mask weights is depicted with a uniform color palette, reflecting their random starting points. As training proceeds, a clear transition in the mask weights is observed: they begin to diverge, exhibiting a pronounced contrast between darker and lighter shades. Notably, the darker weights tend to stabilize and remain consistent in the latter stages of training.  This pattern of change provides visual evidence that the masks are finely optimized with the loss function.

\section{Conclusion}
Previous self-explainable GNNs provide built-in explanations while suffering from a subpar performance in prediction. The feedback explanations are ineffectively applied to supervise the training phase in current GNNs. To address the challenges, we introduce a self-explainable and self-supervised graph neural network (SES) that bridges the explainability and prediction of GNNs by two-phase training. A global mask generator in SES is designed to generate reliable instance-level explanations until explainable training is finished, resulting in notable time savings. The parameters of the graph encoder are shared between two phases of SES. The explanations derived in the explainable training phase are utilized as supervisory information with a self-supervised objective loss during the enhanced predictive learning phase.  Extensive experiments demonstrate that SES achieves SOTA explanation quality and significantly improves the prediction accuracy of current GNNs. Our work illustrates that the tasks of prediction and explainability can be concurrently enhanced during the training of GNNs.

\section*{Acknowledgment}
This work was supported by the National Natural Science Foundation of China (Grant No. 71971002).

\bibliographystyle{unsrt}  
\bibliography{references}

\begin{thebibliography}{10}

\bibitem{metagnn2019}
Fan Zhou, Chengtai Cao, Kunpeng Zhang, Goce Trajcevski, Ting Zhong, and Ji~Geng.
\newblock Meta-gnn: On few-shot node classification in graph meta-learning.
\newblock In {\em CIKM}, pages 2357--2360, 2019.

\bibitem{splitgnn2023}
AK~Awasthi, Arun~Kumar Garov, Minakshi Sharma, and Mrigank Sinha.
\newblock Gnn model based on node classification forecasting in social network.
\newblock In {\em AISC}, pages 1039--1043. IEEE, 2023.

\bibitem{knowledge2021}
Yu~Zhao, Han Zhou, Ruobing Xie, Fuzhen Zhuang, Qing Li, and Ji~Liu.
\newblock Incorporating global information in local attention for knowledge representation learning.
\newblock In {\em IJCNLP}, pages 1341--1351, 2021.

\bibitem{knowledgerepre2023}
Qihuang Zhong, Liang Ding, Juhua Liu, Bo~Du, Hua Jin, and Dacheng Tao.
\newblock Knowledge graph augmented network towards multiview representation learning for aspect-based sentiment analysis.
\newblock {\em IEEE Transactions on Knowledge and Data Engineering}, pages 1--14, 2023.

\bibitem{molecular2018}
Jiaxuan You, Bowen Liu, Zhitao Ying, Vijay Pande, and Jure Leskovec.
\newblock Graph convolutional policy network for goal-directed molecular graph generation.
\newblock In {\em NeurIPS}, volume~31, 2018.

\bibitem{modular2022}
Philip Hawkins, Frederic Maire, Simon Denman, and Mahsa Baktashmotlagh.
\newblock Modular construction planning using graph neural network heuristic search.
\newblock In {\em AJCAI}, pages 228--239. Springer, 2022.

\bibitem{traffic2021}
Mengzhang Li and Zhanxing Zhu.
\newblock Spatial-temporal fusion graph neural networks for traffic flow forecasting.
\newblock In {\em AAAI}, volume~35, pages 4189--4196, 2021.

\bibitem{traffic2022}
Lijun Zhong, Jinjiang Tang, Chen Xu, Baoping Ren, Bin Du, and Zhenyu Huang.
\newblock Traffic prediction of converged network for smart gird based on gnn and lstm.
\newblock In {\em ICBAIE}, pages 341--348. IEEE, 2022.

\bibitem{recsys2018}
Rex Ying, Ruining He, Kaifeng Chen, Pong Eksombatchai, William~L Hamilton, and Jure Leskovec.
\newblock Graph convolutional neural networks for web-scale recommender systems.
\newblock In {\em KDD}, pages 974--983, 2018.

\bibitem{recsys2023}
Wei Wang, Zhenzhen Quan, Siwen Zhao, Guoqiang Sun, Yujun Li, Xianye Ben, and Jianli Zhao.
\newblock User-context collaboration and tensor factorization for gnn-based social recommendation.
\newblock {\em IEEE Transactions on Network Science and Engineering}, pages 1--12, 2023.

\bibitem{senanalys2022}
Shanliang Yang, Linlin Xing, Yongming Li, and Zheng Chang.
\newblock Implicit sentiment analysis based on graph attention neural network.
\newblock {\em Engineering Reports}, 4(1):e12452, 2022.

\bibitem{senanalys2_2022}
Zhigang Jin, Manyue Tao, Xiaofang Zhao, and Yi~Hu.
\newblock Social media sentiment analysis based on dependency graph and co-occurrence graph.
\newblock {\em Cognitive Computation}, 14(3):1039--1054, 2022.

\bibitem{action2021}
Zhan Chen, Sicheng Li, Bing Yang, Qinghan Li, and Hong Liu.
\newblock Multi-scale spatial temporal graph convolutional network for skeleton-based action recognition.
\newblock In {\em AAAI}, volume~35, pages 1113--1122, 2021.

\bibitem{posest2023}
Negar Nejatishahidin, Will Hutchcroft, Manjunath Narayana, Ivaylo Boyadzhiev, Yuguang Li, Naji Khosravan, Jana Ko{\v{s}}eck{\'a}, and Sing~Bing Kang.
\newblock Graph-covis: Gnn-based multi-view panorama global pose estimation.
\newblock In {\em ICCV}, pages 6458--6467, 2023.

\bibitem{textc2019}
Liang Yao, Chengsheng Mao, and Yuan Luo.
\newblock Graph convolutional networks for text classification.
\newblock In {\em AAAI}, volume~33, pages 7370--7377, 2019.

\bibitem{textc2022}
Kunze Wang, Soyeon~Caren Han, and Josiah Poon.
\newblock Induct-gcn: Inductive graph convolutional networks for text classification.
\newblock In {\em ICPR}, pages 1243--1249. IEEE, 2022.

\bibitem{gcn2017}
Thomas~N Kipf and Max Welling.
\newblock Semi-supervised classification with graph convolutional networks.
\newblock In {\em ICLR}, 2017.

\bibitem{gat2018}
Petar Velickovic, Guillem Cucurull, Arantxa Casanova, Adriana Romero, Pietro Lio, and Yoshua Bengio.
\newblock Graph attention networks.
\newblock In {\em ICLR}, 2018.

\bibitem{graphsage2017}
Will Hamilton, Zhitao Ying, and Jure Leskovec.
\newblock Inductive representation learning on large graphs.
\newblock In {\em NeurIPS}, volume~30, page 1025–1035, 2017.

\bibitem{gin2018}
Keyulu Xu, Weihua Hu, Jure Leskovec, and Stefanie Jegelka.
\newblock How powerful are graph neural networks?
\newblock In {\em ICLR}, 2018.

\bibitem{ARMA2021}
Filippo~Maria Bianchi, Daniele Grattarola, Lorenzo Livi, and Cesare Alippi.
\newblock Graph neural networks with convolutional arma filters.
\newblock {\em IEEE Transactions on Pattern Analysis and Machine Intelligence}, 44(7):3496--3507, 2021.

\bibitem{transformconv2021}
Yunsheng Shi, Zhengjie Huang, Shikun Feng, Hui Zhong, Wenjin Wang, and Yu~Sun.
\newblock Masked label prediction: Unified message passing model for semi-supervised classification.
\newblock In {\em IJCAI}, pages 1548--1554, 2021.

\bibitem{fusedgat2022}
Hengrui Zhang, Zhongming Yu, Guohao Dai, Guyue Huang, Yufei Ding, Yuan Xie, and Yu~Wang.
\newblock Understanding gnn computational graph: A coordinated computation, io, and memory perspective.
\newblock In {\em MLSys}, volume~4, pages 467--484, 2022.

\bibitem{anti-symmetric2023}
Alessio Gravina, Davide Bacciu, and Claudio Gallicchio.
\newblock Anti-symmetric {DGN}: a stable architecture for deep graph networks.
\newblock In {\em ICLR}, 2023.

\bibitem{posthoc2010}
Galit Shmueli.
\newblock To explain or to predict?
\newblock {\em Statistical Science}, 25(3):289--310, 2010.

\bibitem{gnnexplainer2019}
Zhitao Ying, Dylan Bourgeois, Jiaxuan You, Marinka Zitnik, and Jure Leskovec.
\newblock Gnnexplainer: Generating explanations for graph neural networks.
\newblock In {\em NeurIPS}, volume~32, pages 9244--9255, 2019.

\bibitem{pgexplainer2020}
Dongsheng Luo, Wei Cheng, Dongkuan Xu, Wenchao Yu, Bo~Zong, Haifeng Chen, and Xiang Zhang.
\newblock Parameterized explainer for graph neural network.
\newblock In {\em NeurIPS}, volume~33, pages 19620--19631, 2020.

\bibitem{pgmexplainer2020}
Minh Vu and My~T Thai.
\newblock Pgm-explainer: Probabilistic graphical model explanations for graph neural networks.
\newblock In {\em NeurIPS}, volume~33, pages 12225--12235, 2020.

\bibitem{graphlime2022}
Qiang Huang, Makoto Yamada, Yuan Tian, Dinesh Singh, and Yi~Chang.
\newblock Graphlime: Local interpretable model explanations for graph neural networks.
\newblock {\em IEEE Transactions on Knowledge and Data Engineering}, 35(7):6968--6972, 2022.

\bibitem{xgnn2020}
Hao Yuan, Jiliang Tang, Xia Hu, and Shuiwang Ji.
\newblock Xgnn: Towards model-level explanations of graph neural networks.
\newblock In {\em KDD}, pages 430--438, 2020.

\bibitem{page2022}
Yong-Min Shin, Sun-Woo Kim, Eun-Bi Yoon, and Won-Yong Shin.
\newblock Prototype-based explanations for graph neural networks (student abstract).
\newblock In {\em AAAI}, volume~36, pages 13047--13048, 2022.

\bibitem{gnninterpreter2023}
Xiaoqi Wang and Han~Wei Shen.
\newblock Gnninterpreter: A probabilistic generative model-level explanation for graph neural networks.
\newblock In {\em ICLR}, 2023.

\bibitem{segnn2021}
Enyan Dai and Suhang Wang.
\newblock Towards self-explainable graph neural network.
\newblock In {\em CIKM}, pages 302--311, 2021.

\bibitem{protgnn2022}
Zaixi Zhang, Qi~Liu, Hao Wang, Chengqiang Lu, and Cheekong Lee.
\newblock Protgnn: Towards self-explaining graph neural networks.
\newblock In {\em AAAI}, volume~36, pages 9127--9135, 2022.

\bibitem{pxgnn2022}
Enyan Dai and Suhang Wang.
\newblock Towards prototype-based self-explainable graph neural network.
\newblock {\em arXiv preprint arXiv:2210.01974}, 2022.

\bibitem{gnnsurvey2021}
Babatounde~Moctard Oloulade, Jianliang Gao, Jiamin Chen, Tengfei Lyu, and Raeed Al-Sabri.
\newblock Graph neural architecture search: A survey.
\newblock {\em Tsinghua Science and Technology}, 27(4):692--708, 2021.

\bibitem{rahg2023}
Kunhao Li, Zhenhua Huang, and Zhaohong Jia.
\newblock {RAHG}: A role-aware hypergraph neural network for node classification in graphs.
\newblock {\em IEEE Transactions on Network Science and Engineering}, 10(4):2098--2108, 2023.

\bibitem{gnnexplainsurvey2022}
Hao Yuan, Haiyang Yu, Shurui Gui, and Shuiwang Ji.
\newblock Explainability in graph neural networks: A taxonomic survey.
\newblock {\em IEEE Transactions on Pattern Analysis and Machine Intelligence}, 45(5):5782--5799, 2022.

\bibitem{self_supervised_survey2020}
Ashish Jaiswal, Ashwin~Ramesh Babu, Mohammad~Zaki Zadeh, Debapriya Banerjee, and Fillia Makedon.
\newblock A survey on contrastive self-supervised learning.
\newblock {\em Technologies}, 9(1):2, 2020.

\bibitem{graphsupervised2022}
Yixin Liu, Ming Jin, Shirui Pan, Chuan Zhou, Yu~Zheng, Feng Xia, and S~Yu Philip.
\newblock Graph self-supervised learning: A survey.
\newblock {\em IEEE Transactions on Knowledge and Data Engineering}, 35(6):5879--5900, 2022.

\bibitem{graphcl2020}
Yuning You, Tianlong Chen, Yongduo Sui, Ting Chen, Zhangyang Wang, and Yang Shen.
\newblock Graph contrastive learning with augmentations.
\newblock In {\em NeurIPS}, volume~33, pages 5812--5823, 2020.

\bibitem{gcc2020}
Jiezhong Qiu, Qibin Chen, Yuxiao Dong, Jing Zhang, Hongxia Yang, Ming Ding, Kuansan Wang, and Jie Tang.
\newblock Gcc: Graph contrastive coding for graph neural network pre-training.
\newblock In {\em KDD}, pages 1150--1160, 2020.

\bibitem{merit2021}
Ming Jin, Yizhen Zheng, Yuan-Fang Li, Chen Gong, Chuan Zhou, and Shirui Pan.
\newblock Multi-scale contrastive siamese networks for self-supervised graph representation learning.
\newblock In {\em IJCAI}, pages 1477--1483, 2021.

\bibitem{heco2021}
Xiao Wang, Nian Liu, Hui Han, and Chuan Shi.
\newblock Self-supervised heterogeneous graph neural network with co-contrastive learning.
\newblock In {\em KDD}, pages 1726--1736, 2021.

\bibitem{molclr2022}
Yuyang Wang, Jianren Wang, Zhonglin Cao, and Amir Barati~Farimani.
\newblock Molecular contrastive learning of representations via graph neural networks.
\newblock {\em Nature Machine Intelligence}, 4(3):279--287, 2022.

\bibitem{messagepassing2017}
Justin Gilmer, Samuel~S. Schoenholz, Patrick~F. Riley, Oriol Vinyals, and George~E. Dahl.
\newblock Neural message passing for quantum chemistry.
\newblock In {\em ICML}, page 1263–1272. PMLR, 2017.

\bibitem{xavier2010}
Xavier Glorot and Yoshua Bengio.
\newblock Understanding the difficulty of training deep feedforward neural networks.
\newblock In {\em AISTATS}, pages 249--256. JMLR Workshop and Conference Proceedings, 2010.

\bibitem{plaintoid2016}
Zhilin Yang, William Cohen, and Ruslan Salakhudinov.
\newblock Revisiting semi-supervised learning with graph embeddings.
\newblock In {\em ICML}, pages 40--48. PMLR, 2016.

\bibitem{polblogs2005}
Lada~A Adamic and Natalie Glance.
\newblock The political blogosphere and the 2004 us election: divided they blog.
\newblock In {\em LinkKDD}, pages 36--43, 2005.

\bibitem{2018pitfall}
Oleksandr Shchur, Maximilian Mumme, Aleksandar Bojchevski, and Stephan G{\"u}nnemann.
\newblock Pitfalls of graph neural network evaluation.
\newblock In {\em Relational Representation Learning Workshop, NeurIPS 2018}, 2018.

\bibitem{lime2016}
Marco~Tulio Ribeiro, Sameer Singh, and Carlos Guestrin.
\newblock " why should i trust you?" explaining the predictions of any classifier.
\newblock In {\em KDD}, pages 1135--1144, 2016.

\bibitem{gnnbaselines2022}
Kai Guo, Kaixiong Zhou, Xia Hu, Yu~Li, Yi~Chang, and Xin Wang.
\newblock Orthogonal graph neural networks.
\newblock In {\em AAAI}, volume~36, pages 3996--4004, 2022.

\bibitem{fidelity+2019}
Phillip~E Pope, Soheil Kolouri, Mohammad Rostami, Charles~E Martin, and Heiko Hoffmann.
\newblock Explainability methods for graph convolutional neural networks.
\newblock In {\em CVPR}, pages 10772--10781, 2019.

\bibitem{silhouettes1987}
Peter~J Rousseeuw.
\newblock Silhouettes: a graphical aid to the interpretation and validation of cluster analysis.
\newblock {\em Journal of {C}omputational and {A}pplied {M}athematics}, 20:53--65, 1987.

\bibitem{caha1974}
Tadeusz Cali{\'n}ski and Jerzy Harabasz.
\newblock A dendrite method for cluster analysis.
\newblock {\em Communications in Statistics-theory and Methods}, 3(1):1--27, 1974.

\bibitem{tsne2008}
Laurens Van~der Maaten and Geoffrey Hinton.
\newblock Visualizing data using t-sne.
\newblock {\em Journal of {M}achine {L}earning {R}esearch}, 9(11), 2008.

\end{thebibliography}

\end{document}